\RequirePackage{fix-cm}
\documentclass[twocolumn]{svjour3}           % twocolumn
\smartqed  % flush right qed marks, e.g. at end of proof
\usepackage{graphicx}%
\usepackage{multirow}%
\usepackage{amsmath,amssymb,amsfonts}%
\usepackage{mathrsfs}%
\usepackage[title]{appendix}%
\usepackage{xcolor}%
\usepackage{textcomp}%
\usepackage{manyfoot}%
\usepackage{booktabs}%
\usepackage{algorithm}%
\usepackage{algorithmicx}%
\usepackage{algpseudocode}%
\usepackage{listings}%
\usepackage[skins]{tcolorbox}
\usepackage[numbers]{natbib}
\usepackage{fbox}
\usepackage{makecell}
\usepackage{inconsolata}
\usepackage{colortbl}
\usepackage{threeparttable}
\usepackage{footnote}
\usepackage{tablefootnote}
\usepackage{subfig,float}
\usepackage{color}
\usepackage[misc]{ifsym}
\usepackage{soul}
\usepackage{titlesec}
\usepackage{pifont}
\usepackage{utfsym}
 \usepackage{cutwin}
\usepackage{multicol}
\newcommand{\revise}[1]{{#1}}

\definecolor{darkergreen}{RGB}{21,152,56}
\definecolor{blue1}{RGB}{11,93,139}
\definecolor{yellow1}{RGB}{90,90,90}
\definecolor{skyblue}{RGB}{0,82,202}
\definecolor{dodgerblue}{RGB}{30,144,255}
\definecolor{lightblue}{RGB}{229,248,255}
\definecolor{green1}{rgb}{0.25,0.5,0.5}
\definecolor{red1}{rgb}{0.7,0.25,0.25}
\definecolor{gray1}{rgb}{0.85,0.85,0.85}
\definecolor{cvprblue}{rgb}{0.21,0.49,0.74}
\definecolor{colred}{RGB}{240,240,240}

\makeatletter
\def\whline#1{%
	\noalign{\ifnum0=`}\fi\hrule \@height #1 \futurelet
	\reserved@a\@xhline}

\usepackage[colorlinks=true,linkcolor=skyblue,filecolor=skyblue,citecolor=skyblue]{hyperref}
\usepackage{geometry}
\begin{document}

\title{
\renewcommand{\windowpagestuff}{
\quad\includegraphics[width=0.8cm]{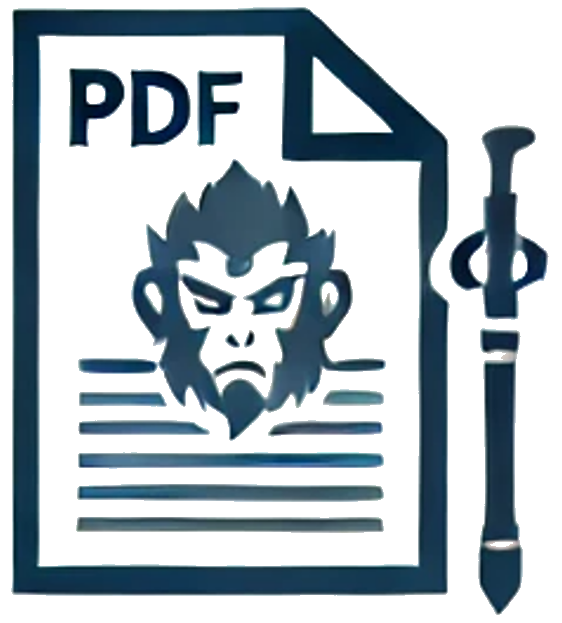}
}
\begin{cutout}{0}{4.0cm}{15cm}{1}
{\color{white} empty} \protect\linebreak
\protect\linebreak
PDF-WuKong\ \ \ : A Large Multimodal Model for Efficient Long PDF Reading with End-to-End Sparse Sampling
\end{cutout}}

\titlerunning{PDF-WuKong: A Large Multimodal Model for Efficient Long PDF Reading with End-to-End Sparse Sampling}

%\twocolumn[

\author{
	Xudong Xie\textsuperscript{1,2}$^*$ \and 
    Hao Yan\textsuperscript{1}$^*$ \and
    Liang Yin\textsuperscript{1}$^*$ \and
    Yang Liu\textsuperscript{1}$^*$ \and 
    Jing Ding\textsuperscript{1} \and
    Minghui Liao\textsuperscript{2}$^\dagger$ \and
    Yuliang Liu\textsuperscript{1}\textsuperscript{\Letter} \and
    Wei Chen\textsuperscript{1}\textsuperscript{\Letter} \and
    Xiang Bai\textsuperscript{1}
}


\institute{
\textsuperscript{1}{Huazhong University of Science and Technology}\\
\textsuperscript{2}{Huawei Inc.} \\
$^*$Equal contribution \\ 
$^\dagger$Project lead \\
{\Letter} Corresponding authors, (ylliu, lemuria\_chen)@hust.edu.cn \\ 
}

\date{Received: 02 Apr 2025 / Accepted: 17 Apr 2026}

\maketitle

\begin{abstract}
Multimodal document understanding is a challenging task to process and comprehend large amounts of textual and visual information. Recent advances in Large Language Models (LLMs) have significantly improved the performance of this task. However, existing methods typically focus on either plain text or a limited number of document images, struggling to handle long PDF documents with interleaved text and images, especially for academic papers. In this paper, we introduce \textbf{PDF-WuKong}, a multimodal large language model (MLLM) that is designed to enhance multimodal question-answering (QA) for long PDF documents. PDF-WuKong incorporates a sparse sampler that operates on both text and image representations, significantly improving the efficiency and capability of the MLLM. \revise{The sparse sampler selects the paragraphs or diagrams most pertinent to user queries.}
To effectively train and evaluate our model, we construct \textbf{PaperPDF}, a dataset consisting of a broad collection of English and Chinese academic papers. Multiple strategies are proposed to build high-quality \textbf{1.1 million} QA pairs along with their corresponding evidence sources. 
Experimental results demonstrate the superiority and high efficiency of our approach over other models on the task of long multimodal document understanding, surpassing proprietary products by an average of \textbf{8.6\%} on F1. Our code and dataset will be released at \href{https://github.com/yh-hust/PDF-Wukong}{https://github.com/yh-hust/PDF-Wukong}.

\end{abstract}

\section{Introduction}

The advent of Large Language Models (LLMs) has significantly advanced the field of PDF document understanding~\cite{saad2023pdftriage,jacob2024constructing,zhang2025smaller}, where these models have demonstrated impressive capabilities in processing and generating human-like text. However, they still face many challenges when it comes to lengthy PDF documents with interlaced text and images, such as academic papers.

\begin{figure}[t]
\centering
\includegraphics[width=0.48\textwidth]{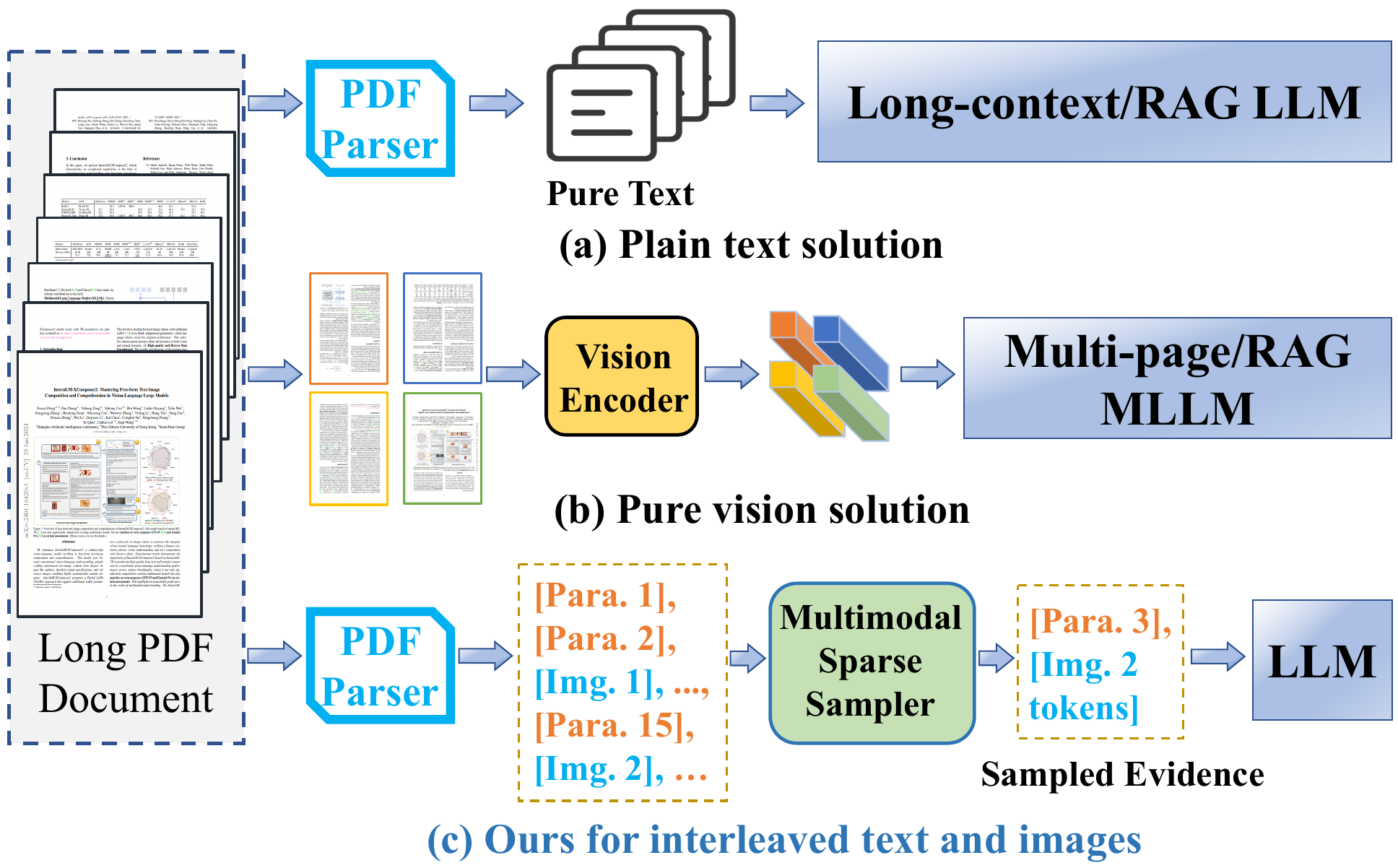}
\caption{Method comparison for long multi-page PDF document understanding. (a) Plain text solution: long-context LLMs and TextRAG-based LLMs for parsed pure text content. (b) Pure vision solution: Multi-page MLLMs with page-level feature interaction and VisualRAG-based MLLMs for page images. (c) Our method is based on end-to-end multimodal sparse sampling for long PDFs with interleaved text and images. \revise{Note: ``LLM'' in (c) is the language backbone component integrated within MLLM.}}
\label{fig:intro}
% \vspace{-2ex}
\end{figure}

To handle lengthy documents, current research in multimodal document understanding with LLMs primarily follows two mainstream technical routes. The first route is based on \emph{pure text modality understanding}. As shown in Fig.~\ref{fig:intro}(a), these approaches typically consider the parsed OCR results from PDF documents, and convert all visual elements into textual representations (e.g., captions and OCR content extracted from figures and tables). They then employ long-context LLMs~\cite{peng2024yarn,chen2024longlora,NEURIPS2023_8511d06d} or utilize Text Retrieval Augmented Generation (RAG) techniques~\cite{yue2023disc,tang2024multihop,edge2024local} to process the textual content. The main disadvantage of this approach is the significant loss of visual information inherent in multimodal documents, making it challenging to support answering fine-grained visual-related questions. 

The second route is based on \emph{pure visual modality understanding}. As illustrated in Fig.\ref{fig:intro}(b), this approach generally avoids parsing PDF documents and instead treats each page as an image, and utilizing multimodal LLMs for document understanding tasks~\cite{liu2024textmonkey,chen2024far,wei2025vary}. However, high-resolution images and numerous pages generate a large number of visual tokens, significantly increasing the models' input token length and leading to scalability issues. This makes it challenging to process multi-page long documents efficiently. While some recent multi-page visual document understanding models handle up to 8 pages~\cite{liu2024focus} or 20 pages~\cite{TITO2023109834}, they typically encode each page separately and perform page-level visual feature interactions such as concatenation~\cite{SlideVQA2023,blau2024gram}. Nevertheless, as the number of pages grows, computational resource consumption escalates substantially, rendering these models inefficient for processing longer documents. Although some recent VisualRAG-based models~\cite{yu2024visrag,chen2025svrag} try to find document pages related to the query, they perform poorly for text-intensive scenarios and cannot provide more fine-grained evidence.

Considering the limitations of existing methods in multimodal understanding of long PDF documents, we propose a new MLLM architecture with end-to-end sparse sampling, named \textbf{PDF-WuKong}. Since most user queries are only related to a small part
of the content in a long document, the sparse sampling can significantly remove redundant noise information.
It encodes text paragraphs and diagrams in the parsed PDF document, utilizing both text and image representations to identify and extract the most relevant evidence in response to a user's query. The sampled sparse evidence significantly reduces the number of input tokens of the LLM and this process is independent of the length of the input documents. Moreover, the sparse sampler and LLM can be integrated in an end-to-end manner for training and inference, optimizing the performance of multimodal representation and question answering while improving time efficiency. \revise{It is worth noting that this sparse sampler is a flexible and generic design that can be applied to various MLLM architectures through fine-tuning.} Another important characteristic is that it can naturally provide strong interpretability for question answering.

In order to simultaneously represent and understand the multimodal content of documents and further improve the ability to process long PDF documents, we construct a training dataset specifically for English and Chinese academic paper PDFs. The academic paper PDF is a kind of typical document that contains rich interleaved text and images, which can intuitively reflect the challenges of our task and the advantages of our model. The dataset contains complete PDF documents, professional academic questions, answers, and evidence sources for the answers, based on multiple construction
strategies. We also provide a corresponding bilingual benchmark named \textbf{PaperPDF}.

\setlength{\tabcolsep}{2pt}
\begin{table*}[t!]
% \small
% \vspace{-4ex}
\begin{center}
% \small
\caption{Comparison of various models for processing multi-page long documents.} 
\begin{tabular}{c|c|c|p{9.5cm}}
\toprule
 \textbf{Input modality}  & \textbf{Type}& \textbf{Number of tokens} & \textbf{Models}  \\
% \midrule
\hline
%\cline{5-6} \cline{7-8}
 \multirow{2}{*}{Plain text}  & Long-context & Linear increase & LongLoRA~\cite{chen2024longlora}, LongLLaMA~\cite{NEURIPS2023_8511d06d}, YaRN~\cite{peng2024yarn} \\
 \cline{2-4}
 % \cmidrule(lr){2-4}
&Text RAG &w/o Linear increase & Graph RAG~\cite{edge2024local}, DISC-LawLLM~\cite{yue2023disc}, RAPTOR~\cite{sarthi2024raptor}\\
% \midrule
\hline
 \multirow{2}{*}{Pure vision}  
 
 % \cmidrule(lr){2-4}
 & Multi-page & Linear increase &\revise{ Fox~\cite{liu2024focus}, DocOwl2~\cite{hu2024mplugdocowl2}, MiniCPM-V~\cite{yao2024minicpm,yu2025minicpm}, InternVL~\cite{InternVL2.5,wang2025internvl3_5}, Qwen-VL~\cite{Qwen2.5-VL, Qwen3-VL}, Granite Vision~\cite{team2025granite} }, DocSeeker~\cite{yan2026docseeker}\\
 \cline{2-4}
 & Visual RAG & w/o Linear increase &  VisRAG~\cite{yu2024visrag}, SV-RAG~\cite{chen2025svrag} \\
 % \midrule
 \hline
 \multirow{2}{*}{Text and images}  & \revise{ Multi-page} & \revise{Linear increase} & \revise{Hi-VT5~\cite{TITO2023109834}, GRAM~\cite{blau2024gram} } \\
 \cline{2-4}
 & Multimodal RAG & w/o Linear increase & PDF-WuKong \textbf{(Ours)}  \\

\bottomrule
\end{tabular}

\label{related}
\end{center}

\end{table*}

We train PDF-WuKong on the PaperPDF dataset, complemented by general-domain document question-answering datasets. Experimental results substantiate the effectiveness and efficiency of our approach to the task of long multimodal PDF understanding. PDF-WuKong significantly outperforms potential open-source models that may be applied to this task. It also surpasses some proprietary products for document understanding on our proposed PaperPDF benchmark. As the number of document pages increases, its accuracy and efficiency will not decrease significantly. It also achieves competitive performance on several document-oriented VQA datasets, especially multi-page benchmarks like DUDE~\cite{van2023document}. Besides, for the recent benchmark MM-NIAH~\cite{wang2024needle} of long multimodal documents, PDF-WuKong also outperforms other models with fewer parameters. Our model achieves the best performance on multimodal content with a context length of 64K.

The \textbf{main contributions} of this paper are as follows:

% \setlength{\tabcolsep}{2pt}
% \begin{table*}[t!]

\begin{itemize}
\item We introduce a large multimodal model for long PDF understanding with end-to-end sparse sampling, achieving accurate and efficient performance and providing explainable evidence for the answer.
\item We propose a bilingual PDF multimodal question answering dataset (\textbf{PaperPDF}) with $1.1M$ QA pairs for training and $10k$ QA pairs for evaluation.
\item Our model significantly surpasses existing open-source models and proprietary products (by an average of $8.6\%$ on F1) on long multimodal PDF understanding.
\end{itemize}

\section{Related Works}

\subsection{Document Understanding Datasets}

Earlier document understanding datasets only focused on the NLP tasks such as summarization~\cite{huang2021efficient}, and QA~\cite{dasigi2021dataset} of plain text. 
Meanwhile, there were several visual document datasets mainly aimed at text perception tasks such as Document Layout Analysis (DLA)~\cite{zhong2019publaynet,li2020docbank,pfitzmann2022doclaynet} and Key Information Extraction (KIE)~\cite{simsa2023docile,park2019cord,huang2019icdar2019}. 
% For example, DocBank~\citep{li2020docbank} constructs 500K high-quality document pages to enable the document layout model to utilize both textual and visual information. 
Recently, more datasets have been proposed for multimodal document QA across various scenarios. For example, DocVQA~\cite{Mathew_2021_WACV} and OCRVQA ~\cite{MishraSSC19} provide single-page QA data on books and business documents. ChartQA~\cite{masry2022chartqa} and ChartX~\cite{xia2024chartx} focus on visual reasoning for chart documents. ArXivQA~\cite{li2024multimodal} extracts scientific figure-caption data from the arxiv papers to enhance the academic ability of MLLMs. InfoVQA~\cite{Mathew_2022_WACV} contains many infographic documents that are a combination of textual, graphical and visual elements. 
However, these visual document datasets only define the single-page VQA task.
% and current MLLMs~\cite{Qwen2VL,chen2024far} have achieved remarkable performance on it.

There are also some multi-page QA datasets~\cite{SlideVQA2023,van2023document,TITO2023109834} that require the model to understand the content relationship via multi-hop reasoning and capture the crucial information from multi-page documents. MP-DocVQA~\cite{TITO2023109834} extends DocVQA~\cite{Mathew_2021_WACV} by adding the context pages. DUDE~\cite{van2023document} constructs multi-page QA data from multi-industry and multi-domain documents.
Additionally, a large-scale dataset designed for the comprehensive understanding of lecture slides has been introduced~\cite{zhang2025towards}.
DocGenome~\cite{xia2024docgenome} was constructed as a scientific document benchmark. MM-NIAH~\cite{wang2024needle} is a benchmark evaluating the capability of MLLMs to comprehend long multimodal documents, which requires the model to answer according to the key information scattered throughout the document. 
However, the answers in these datasets lack evidence and cannot provide reliable interpretability, especially for questions that require referring to multiple pieces of evidence in long documents.

\subsection{Document Understanding Methods}

Existing document understanding methods typically focus on either plain text or a limited number of document images.
Methods that rely on pure text modality aim to process documents by first converting them into plain text through document parsing or Optical Character Recognition (OCR)~\cite{zhao2024cbnet,zheng2024cdistnet,wu2024end,liu2025multi}. They then employ efficient long-context mechanisms to handle long texts, such as sparse attention~\cite{chen2024longlora}, memory networks~\cite{NEURIPS2023_8511d06d}, or position interpolation~\cite{peng2024yarn}.  Besides, the methods based on retrieval-augmented generation (RAG)~\cite{yue2023disc,sarthi2024raptor,edge2024local} also show impressive capabilities for long texts. 
While these approaches can integrate visual elements by including image captions or transforming images into natural language descriptions, they struggle with a fine-grained understanding of visual information. This restricts their effectiveness in tasks requiring detailed interpretation of textual and visual components within documents.

Another solution is visual document understanding with purely visual input, treating each document page as an image. 
 Many MLLMs such as UniDoc~\cite{feng2023unidocuniversallargemultimodal}, mPLUG-DocOwl~\cite{ye2023mplugdocowl}, StrucTexTv3~\cite{lyu2024structextv3} and Vary~\cite{wei2025vary} can perform this task in an OCR-free manner. 
 % Vary~\cite{wei2025vary} employs an extra SAM-style~\cite{SAM} vision vocabulary specific to document and chart data, enabling direct encoding of entire pages with high compression ratios.
 Other researchers have advanced the understanding of high-resolution document pages by dividing input images into smaller patches, such as UReader~\cite{ye2023ureader}, TextMonkey~\cite{liu2024textmonkey}, and LLaVA-NeXT~\cite{liu2024llavanext}. DocPedia~\cite{feng2023docpedia} processes high-resolution images in the frequency domain via the DCT transformation.
 % InternLM-XComposer2-4KHD~\cite{internlmxcomposer2_4khd} and 
 InternVL-V1.5~\cite{chen2024far} introduces a dynamic
resolution mechanism to capture more details. 
The reliance on high resolution results in a higher number of tokens and cannot be extended to multi-page documents. Recently, CoPali~\cite{faysse2025colpali} uses Vision LLM to retrieve query-related image patches in a single-page image, but it has not been applied to multi-page documents and only conducts the retrieval task without the QA task.

\revise{There are several visual models specifically designed for multi-page documents. Fox~\cite{liu2024focus} unifies all image tokens of up to 8 pages into a sequence to achieve multi-page QA. mPLUG-DocOwl2~\cite{hu2024mplugdocowl2} compresses each high-resolution document image into 324 tokens, guided by low-resolution global visual features. Notably, DocSeeker~\cite{yan2026docseeker} embeds the retrieval capability directly into the model rather than relying on a separate RAG pipeline, introducing structured visual reasoning to achieve accurate and fine-grained evidence grounding in multi-page documents. To improve model interpretability, DocThinker~\cite{yu2025docthinker} leverages rule-based reinforcement learning for explainable document understanding.
Besides, some general MLLMs such as MiniCPM-V~\cite{yao2024minicpm,yu2025minicpm}, InternVL~\cite{InternVL2.5,wang2025internvl3_5}, Qwen-VL~\cite{Qwen2.5-VL, Qwen3-VL}, and Granite Vision~\cite{team2025granite} can already support multi-image input by training with multi-image samples.}
 % However, these models are limited in the length of documents they can process at one time; as document length increases, inference time and memory usage grow significantly, hindering their ability to efficiently capture rich visual content in lengthy documents.
These models encode each page image separately and then perform page-level visual feature interactions such as concatenation.
However, more pages generate more visual tokens, which
brings greater resource consumption to the LLM. This makes them inefficient and unable to process longer documents. 

Recently, some document understanding MLLMs based on pure visual RAG~\cite{yu2024visrag,chen2025svrag,yinmstar} have been proposed. They usually separate the page image retrieval and document understanding generation into two independent stages during training and inference. They still treat each page as a separate image and use pure visual retrieval to obtain page-level evidence. These retrieval schemes remain challenging for text-intensive document understanding due to insufficient text recognition capabilities. Moreover, they can only obtain page-level evidence rather than more fine-grained evidence.

\revise{Beyond plain text and pure vision approaches, certain specialized models for multi-page documents attempt to integrate multimodal information by concatenating OCR tokens with visual features for each page. Hi-VT5~\cite{TITO2023109834} and GRAM~\cite{blau2024gram} are representative works in this category. They combine tokens of multiple pages through hierarchical transformer architecture~\cite{TITO2023109834} or global-local reasoning~\cite{blau2024gram}, being able to handle up to 20 pages. However, their token count still increases linearly with the number of pages.}

Therefore, we propose first to parse the lengthy document into interleaved text and image content and then perform multimodal sparse sampling in an end-to-end manner. Tab.~\ref{related} summarizes the attributes of each type of method.

\section{Methodology}

\begin{figure*}
\centering
\includegraphics[width=0.9\textwidth]{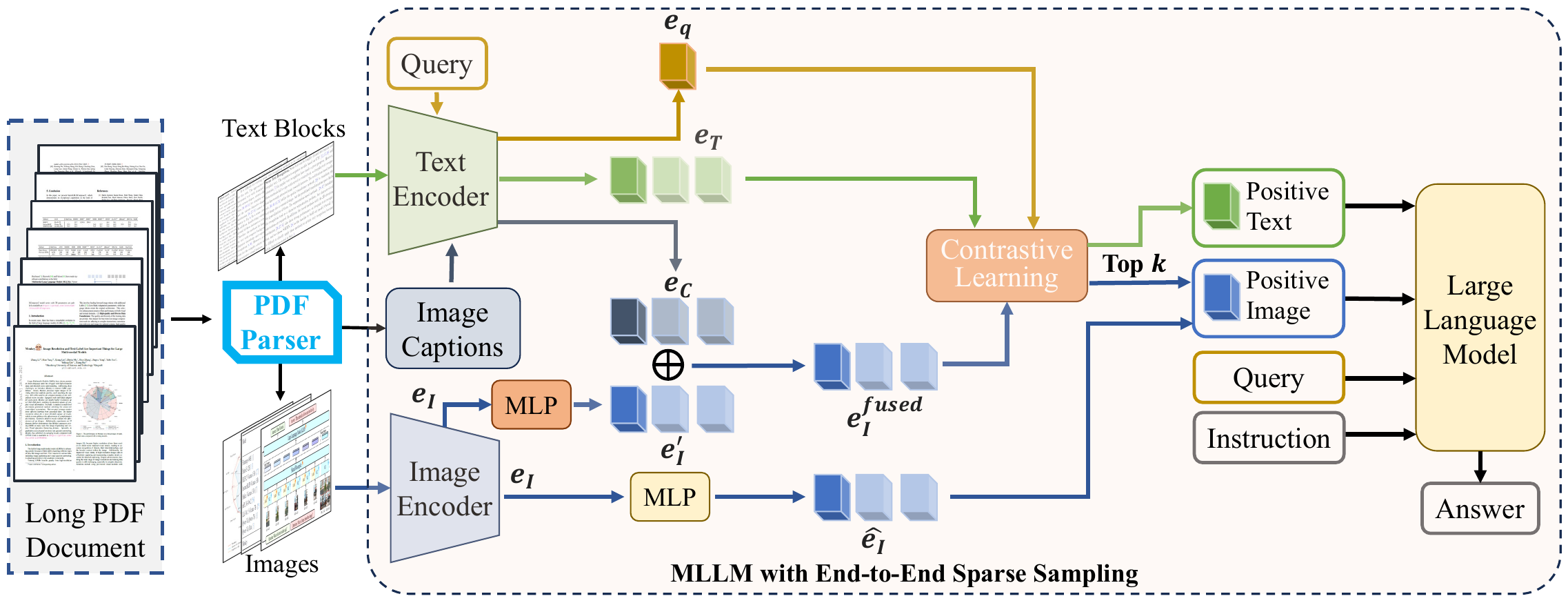}
\caption{The overall structure of PDF-WuKong consists of a document parser, a multimodal sparse sampler that integrates end-to-end with an LLM. \revise{``LLM'' in the figure refers to the language backbone within the MLLM.}}
\label{fig:pipeline}
\end{figure*}

\subsection{Motivation and Method Overview}

% Our pipeline consists of three components: a document parser, a sparse sampler, and a large language model, as shown in Fig.~\ref{fig:pipeline}. The document parsing stage converts input PDFs into machine-readable content with interleaved text and images. The sparse sampler then encodes and caches embeddings for text blocks and images separately. When receiving a user query, it retrieves the most relevant content through similarity matching. Finally, the query, the sampled raw text blocks, and the sampled image tokens are fed into the LLM for answer generation. The detailed procedure is outlined in Algorithm~\ref{alg:pipeline}.

The central motivation behind PDF-WuKong is to empower multimodal large language models (MLLMs) to efficiently deal with sophisticated multimodal reasoning and comprehension tasks on lengthy multimodal PDF documents, particularly academic papers densely populated with paragraphs, figures, and tables. Directly inputting entire documents, including all textual paragraphs, figures, and tables, into existing MLLMs is computationally prohibitive due to their substantial length. Furthermore, crucial information relevant to a user's query may be diluted by irrelevant content, making it challenging for the model to accurately locate and reason about key evidence. 

To address these challenges, our overarching insight is to leverage sparse multimodal sampling to selectively identify and retrieve the most relevant text and image blocks given a user query. Crucially, instead of introducing a fully independent multimodal retrieval model, we propose a lightweight retrieval strategy that reuses the visual encoder from the pretrained MLLM. This strategy reduces computational redundancy and facilitates efficient embedding caching, enabling fast inference and effective end-to-end training.

As illustrated in Fig.~\ref{fig:pipeline}, PDF-WuKong comprises three main components: 1) \emph{Document Parsing} converts PDF documents into structured, interleaved sequences of text and image blocks (Sec.~\ref{sec:parsing}); 2) \emph{Multimodal Sparse Sampling} encodes and caches compact embeddings for retrieval, efficiently identifying query-relevant sparse subsets of blocks (Sec.~\ref{sec:sparse_sampling}); 3) \emph{Multimodal Response Generation} takes the retrieved sparse set of multimodal blocks and synthesizes concise, accurate responses to user queries using a pretrained MLLM (Sec.~\ref{sec:generation}).
The overall inference process is summarized in Algorithm~\ref{alg:pipeline}. 
% and the end-to-end training procedure is detailed in Algorithm~\ref{alg:training}.

\subsection{Document Parsing and Representation}
\label{sec:parsing}

Given a PDF document \(D\), we first parse it into a structured sequence of multimodal blocks, explicitly preserving its layout and reading order. Specifically, we extract:

- A set of \emph{text blocks} \(\{T_1, T_2, \dots, T_n\}\), each typically corresponding to a paragraph.

- A set of \emph{image blocks} \(\{I_1, I_2, \dots, I_m\}\), each including figures or tables with their captions. 

If an image block has an accompanying textual caption, we explicitly associate them together. The structured multimodal content is stored as an XML file and parsed offline using open-source PDF parsing tools Grobid~\cite{GROBID} and MinerU~\cite{wang2024mineruopensourcesolutionprecise}. 
% This pre-processing step greatly accelerates inference by enabling offline caching of embeddings.

\revise{
To ensure robust parsing of complex academic layouts, we employ a hybrid parsing pipeline combining Grobid~\cite{GROBID} for metadata extraction and MinerU~\cite{wang2024mineruopensourcesolutionprecise} for fine-grained layout analysis. We selected this specific combination after comparing multiple open-source alternatives.
First, both tools are lightweight and computationally efficient, enabling the scalable processing of our massive 1.1M document corpus. Second, MinerU is specifically optimized for academic documents; it demonstrates superior performance in parsing complex elements such as mathematical formulas and charts compared to traditional rule-based parsers. Crucially, its robust support for Chinese character recognition ensures high-quality parsing for the Chinese subset of our bilingual dataset. This pipeline effectively mitigates common parsing errors in mixed-content documents. }

\subsection{Multimodal Sparse Sampling}
\label{sec:sparse_sampling}

To effectively handle lengthy documents, our sparse sampling approach aims to identify a concise subset of the most query-relevant text and image blocks. To achieve this, we encode each block into a compact embedding vector, enabling efficient retrieval through similarity matching in a unified embedding space.

\textbf{Text Embedding.} We introduce a lightweight text encoder \(E_T(\cdot)\) for each text block \(T_i\), resulting in a single text embedding vector \(e_{T_i}\in\mathbb{R}^{d_T}\) that concisely captures semantic content.

\textbf{Image Embedding.} \revise{For each image block \(I_j\), we reuse the visual encoder \(E_V(\cdot)\) from the pretrained MLLM to obtain the visual sequence embeddings \(e_{I_j}\in\mathbb{R}^{n \times d_V}\), where $n$ is the number of visual tokens and $d_v$ is the visual embedding dimension. To unify visual and textual embeddings into a shared embedding space, we aggregate the visual sequence into a single compact vector and project it into the textual embedding space $d_t$ using the projection head ${MLP_{proj}}$: }

\begin{equation}
\revise{e_{I_j}^{\prime} = MLP_{proj}(MeanPool(e_{I_j})), \quad e_{I_j}^{\prime}\in\mathbb{R}^{d_T}.}
\end{equation}

% \begin{equation}
%     e'_{I} = \mathrm{MLP_{proj}}(\mathrm{Pool}(H_I)), \quad e'_{I} \in \mathbb{R}^{D_t}
% \end{equation}

If an image block has an associated textual caption, we further enrich its embedding by encoding the caption text using the same text encoder \(E_T(\cdot)\) and obtain a single caption embedding vector. The final embedding $e_{I_j}^{\text{fused}}$ for the image is obtained by element-wise addition of the projected visual embedding and the caption embedding. 

\textbf{Query-driven Retrieval.} Given a user query \(q\), we also encode it using the text encoder \(E_T(\cdot)\) to obtain a compact query embedding vector in \(e_{q}\in\mathbb{R}^{d_T}\). Then, we can compute cosine similarity between the query embedding and all cached compact embeddings of text blocks (\(\{e_{T_i}\}\)) and image blocks (\(\{e_{I_j}^{\text{fused}}\}\)). The top-\(K\) most similar blocks are selected as sparse evidence for multimodal response generation. 

\textbf{Offline Caching.} All compact embeddings for text and image blocks, along with the original visual sequence embeddings, are computed offline and cached. This offline caching strategy significantly speeds up inference by avoiding redundant visual encoding computations during response generation.

\subsection{Multimodal Response Generation}
\label{sec:generation}

Given the retrieved sparse multimodal blocks and the user query, we generate the final response through the pretrained MLLM. \revise{For image blocks, we reuse the cached visual sequence embeddings $\{e_{I_j}\}$, containing all image patch tokens. Since these are derived from the MLLM's native visual encoder, they can be directly projected into the input space of the MLLM's language backbone using the original connection module $MLP_{conn}$, preserving fine-grained visual details: }
\begin{equation}
\revise{\hat{e_{I_j}} = MLP_{conn}(e_{I_j}), \quad \hat{e_{I_j}}\in\mathbb{R}^{n\times d_{llm}},}
\end{equation}
\revise{$\hat{e_{I_j}}$ serves as the visual context tokens fed into the LLM backbone. This approach eliminates redundant visual encoding operations at inference time.}

\begin{algorithm}[H]
\caption{Inference pipeline for PDF-WuKong}
\label{alg:pipeline}
\begin{algorithmic}[1]

\State \textbf{Input:} PDF document $D$, user query $q$
\State \textbf{Output:} Generated answer $a$
\State \textbf{Initialize:} Text encoder $E_T$, image encoder $E_V$, \text{LLM}, and two MLP projectors $\text{MLP}_{\text{proj}}$, $\text{MLP}_{\text{conn}}$

\State \textbf{Stage 1: Document Parsing}

\State Parse the input document $D$ into text blocks and images with their captions:
\[
    \{T_i\}, \{(I_j, C_j)\} \leftarrow \text{Parser}(D)
\]

\State \textbf{Stage 2: Sparse Sampling}

\State Encode all text blocks and images and \textbf{cache} all candidate vector embeddings:
\[
     \{e_{T_i}\} \leftarrow E_T (\{T_i\}) , 
\]
\[
      \{e_{I_j}\} \leftarrow E_V(\{I_j\}),
\]
\[
     \{e_{C_j}\} \leftarrow E_T (\{C_j\}) , 
\]
\[
\revise{\{e_{I_j}^{\prime}\} \leftarrow \text{MLP}_{\text{proj}}(MeanPool(\{e_{I_j}\})),}
\]
\[
     \{e_{I_j}^{\text{fused}}\} \leftarrow \{e_{I_j}^{\prime}\} + \{e_{C_j}\} 
\]

% \State Cache and store all candidate vector embeddings $\{E_T, E_I\}$.

% \State \textbf{Stage 3: Query Processing}

\State Encode the user query $q$:
\[
    e_q \leftarrow E_T(q)
\]

\State Calculate the similarity between query embedding $e_q$ and cached text/image embeddings $\{e_{T_i}\}$, $\{e_{I_j}^{\text{fused}}\}$:
\[
    S_T = \{\text{sim}(e_q, e_{T_i}) \mid i = 1, 2, \dots, n\},
\]
\[
    S_I = \{\text{sim}(e_q, e_{I_j}^{\text{fused}}) \mid j = 1, 2, \dots, m\}
\]

\State \revise{Select top-$k$ relevant text blocks and images with captions:}
\[
    \revise{(T, I, C)_{top} \leftarrow \text{TopK}(S_T, S_I, k)}
\]

\State \textbf{Stage 3: Answer Generation}

\State Input the query $q$ and the selected top-$k$ text blocks $T_i$ and visual embeddings $e_{I_j}$ into the LLM:
\[
\{\hat{e_{I_j}}\} \leftarrow \text{MLP}_{\text{conn}}(\{e_{I_j}\}),
\]
\[
   \revise{ a \leftarrow \text{LLM}(q, (T_i, \hat{e_{I_j}}, C_i)_{top}) }
\]

\State \textbf{Return} the generated answer $a$.

\end{algorithmic}
\end{algorithm}

\revise{
Conversely, for text blocks, we provide raw textual input. Reusing retrieval embeddings is not feasible because 1) the text encoder (BGE-M3) and the MLLM's language module use different tokenizers and vocabularies, and 2) LLMs are optimized to process discrete token indices mapped to their own internal embedding spaces. Providing raw text ensures the model utilizes its pre-trained linguistic capabilities to the fullest extent.}

\revise{It is worth noting that we utilize the original visual embeddings $\{e_{I_j}\}$ for generation rather than the fused embeddings $\{e_{I_j}^{fused}\}$ used in retrieval. This decision is based on the fact that the $MLP_{conn}$ is pre-trained to align pure visual features with the LLM space; introducing fused embeddings at this stage would cause a distribution mismatch.}

\revise{To encourage transparent reasoning and enhance semantic understanding, we explicitly annotate each input block with metadata such as block indices and section headings. Specifically, for retrieved image blocks, their corresponding captions are also provided as plain text inputs to the LLM alongside the visual embeddings.} The MLLM is trained to explicitly reference relevant block indices when generating responses, thereby enhancing interpretability and evidence grounding.

% \subsection{Joint End-to-End Training Objectives}
% \label{sec:training}

% PDF-WuKong is trained end-to-end using a combination of two complementary learning objectives: (1) a contrastive representation learning loss \(\mathcal{L}_{\text{CTR}}\) for effective embedding alignment, and (2) a cross-entropy loss \(\mathcal{L}_{\text{CE}}\) for accurate multimodal response generation.

% \textbf{Contrastive Representation Loss (\(\mathcal{L}_{\text{CTR}}\)).} For each training instance (query-response pair), positive blocks (\(B^+\)) are those explicitly annotated as relevant evidence, while negative blocks (\(B^-\)) are randomly sampled irrelevant blocks from the document. We optimize embeddings by minimizing the following contrastive loss:

% \begin{equation}
% \mathcal{L}_{\text{CTR}} = - \frac{1}{|B^+|}\sum_{e^+\in B^+}\log\frac{\exp(\text{sim}(E_T(q), e^+)/\tau)}{\exp(\text{sim}(E_T(q), e^+)/\tau)+\sum_{e^-\in B^-}\exp(\text{sim}(E_T(q), e^-)/\tau)},
% \end{equation}

% where \(\text{sim}\) denotes cosine similarity, and \(\tau\) is a temperature hyperparameter.

\subsection{Joint End-to-End Training Objectives}
\label{sec:training}

PDF-WuKong is trained end-to-end using a combination of two complementary learning objectives: (1) a contrastive representation learning loss \(\mathcal{L}_{\text{CTR}}\) for effective embedding alignment, and (2) a cross-entropy loss \(\mathcal{L}_{\text{CE}}\) for accurate multimodal response generation.

\textbf{Contrastive Representation Loss (\(\mathcal{L}_{\text{CTR}}\)).} For each training instance (query-answer pair), positive blocks (\(B^+\)) are those explicitly annotated as relevant evidence, while negative blocks (\(B^-\)) are randomly sampled irrelevant blocks from the document. Given a query embedding \(e_{q}\in\), we optimize embeddings by minimizing the following contrastive loss:

\begin{equation}
\mathcal{L}_{\text{CTR}} = - \frac{1}{|B^+|}\sum_{e^+\in B^+}\log\frac{\exp\bigl(\text{sim}(e_{q}, e^+)/\tau\bigr)}{\mathcal{Z}},
\end{equation}
where the normalization term \(\mathcal{Z}\) sums over positive and negative samples: 

$$\mathcal{Z} = \exp\bigl(\text{sim}(e_{q}, e^+)/\tau\bigr) + \sum_{e^-\in B^-}\exp\bigl(\text{sim}(e_{q}, e^-)/\tau\bigr).$$

Here, \(\text{sim}(\cdot,\cdot)\) denotes cosine similarity, and \(\tau\) is a temperature hyperparameter controlling the softness of the similarity distribution.

\textbf{Cross-Entropy Response Generation Loss (\(\mathcal{L}_{\text{CE}}\)).} The second objective trains the MLLM to generate accurate textual responses conditioned on the retrieved multimodal evidence and the query:
\begin{equation}
\mathcal{L}_{\text{CE}} = -\sum_{l=1}^{L}\log p_\theta(r_l^*|q,B^+,r_{<l}^*),
\end{equation}
where \(r^*\) is the ground-truth response sequence, and \(p_\theta\) is the MLLM.

\textbf{Joint Optimization.} Our end-to-end training objective integrates both losses:
\begin{equation}
\mathcal{L}_{\text{total}} = \mathcal{L}_{\text{CTR}} + \mathcal{L}_{\text{CE}}.
\end{equation}

% This joint strategy enables efficient multimodal embedding alignment and robust multimodal reasoning, effectively addressing computational challenges associated with lengthy multimodal document understanding tasks.

Overall, PDF-WuKong introduces joint alignment to reconcile efficiency and fidelity: the visual features first map to a text-dominated metric space for coarse cross-modal retrieval, then to the LLM-aligned compositional space for fine-grained reasoning. This design mirrors human coarse-to-fine comprehension, filtering salient evidence via unified semantics before synthesizing grounded responses. Meanwhile, it avoids redundant recomputation through parameter-efficient adaptation and offline caching. The duality resolves modality granularity conflicts, ensuring computational tractability without sacrificing pretrained multimodal synergies. 

% By jointly optimizing contrastive evidence grounding and multimodal reasoning, sparse multimodal semantics are dynamically aligned through query-aware attention while preserving pretrained knowledge via parameter-efficient adaptation

% PDF-WuKong achieves cognitive-aligned document understanding by jointly optimizing contrastive evidence grounding and cross-modal reasoning, where sparse multimodal semantics are dynamically aligned through query-aware attention while preserving pretrained knowledge via parameter-efficient adaptation. Its training rationality originates from emulating human analytical workflows—contextual saliency detection followed by focused compositional synthesis—thereby resolving the tension between computational efficiency and holistic comprehension in long-form multimodal reasoning.

% During the training, we input the positive text $T_{P}$ and the positive image tokens $e_{I_P}$ into the LLM. Besides, the query and instruction are also input into the LLM. Then, we calculate the cross-entropy loss $\mathcal{L}_{\text{QA}}$ between the output answer $a$ and the ground truth. 
% Finally, the total optimization objective is:
% \begin{equation}
% \mathcal{L}_{\text{total}} = \mathcal{L}_{\text{rep}} +  \mathcal{L}_{\text{QA}}.
% \end{equation}
% PDF-WuKong is optimized end-to-end by these two loss functions for effective multimodal alignment and QA. The pseudocode for training is shown in Algorithm~\ref{alg:training}.

\section{PaperPDF Dataset}
\label{sec:dataset}

\subsection{Overview}

In order to enable PDF-WuKong to simultaneously represent and understand the multimodal content of documents and provide explainable answer evidence, we construct a dataset with English and Chinese academic paper PDFs. The dataset contains complete PDF documents, professional academic questions, answers, and evidence sources for the answers, based on multiple construction strategies. 
Specifically, our motivation for creating PaperPDF is threefold: (1) In the long document QA task, answers often derive from specific document segments, with other content acting as noise and complicating MLLM reasoning; (2) Existing datasets are either limited to single-page documents or lack fine-grained evidence ground truth, hampering the training of our sparse sampler; (3) There is currently a lack of a bilingual multi-page multimodal document QA dataset. 
Therefore, we introduce multiple strategies to construct question-answer pairs from long documents for training and evaluation.

\begin{figure*}
\centering
\includegraphics[width=1\textwidth]{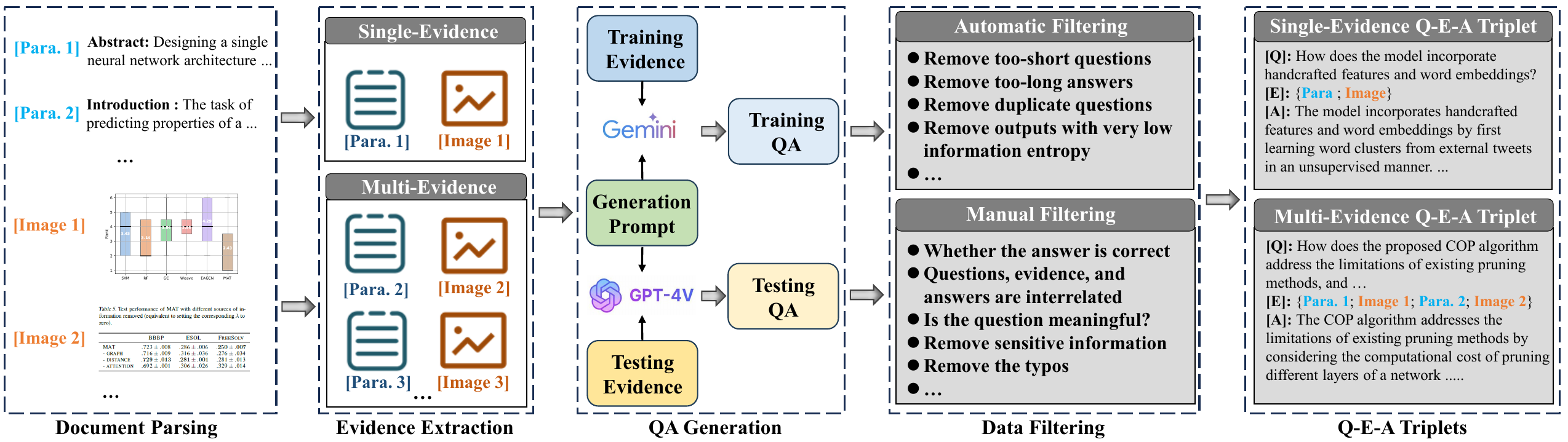}
\caption{The construction process of PaperPDF based on single evidence and multiple evidence.}
%\caption{The construction process of Paper-PDFQA.}
\label{fig:data}
\end{figure*}

\subsection{Dataset construction} 
The academic paper PDF is a kind of typical document that contains rich interleaved text and images, which can intuitively reflect the challenges of our task and the advantages of our model. Therefore, we first obtained approximately 60k English and 10k Chinese documents from open-access academic paper websites, encompassing nearly 70 disciplines, such as computer science, engineering, and materials science.

To build high-quality data, we propose a four-step construction strategy and design a variety of evidence distributions based on human questioning habits. The construction strategy consists of document parsing, evidence extraction, QA generation, and data filtering, as shown in Fig.~\ref{fig:data}. 

\textbf{First}, each PDF is parsed as interleaved text and image chunks according to the statment in Sec.~\ref{sec:parsing}.

\textbf{Second}, we extract evidence from each PDF according to different extraction rules, and the type of evidence in our PaperPDF can be divided into two categories:

Single-evidence extraction. The evidence can be categorized into \textit{Text-only} and \textit{Image-only}. We randomly extract a text paragraph or a diagram from a document and construct questions and answers based only on the information in this single evidence. We stipulate that the length of the extracted paragraphs must contain at least 3 sentences, and the diagram needs to be extracted together with its caption. These enable PDF-WuKong to acquire the initial capabilities of sparse sampling and long multimodal document understanding.

Multi-evidence extraction. The evidence can be categorized into \textit{Image-text}, \textit{Section}, and \textit{Cross-paragraph}. These facilitate reasoning across multiple chunks, involving combinations of text and images. The \textit{Image-text} evidence is derived from a text chunk and the diagrams it refers to. It is also extracted from a diagram and the text paragraphs that reference it.
The \textit{Section} evidence utilizes all chunks in a section, including all the text paragraphs, figures, and tables. The \textit{Cross-paragraph} evidence involves related paragraphs across the document. To extract this type of evidence, we first conduct semantic summarizations for each paragraph and then select several of the most semantically relevant text chunks. These multi-evidence samples can significantly enhance our model's multi-hop reasoning capability. 
% Although more complex, multi-evidence pairs significantly improve sampler accuracy and MLLM multimodal reasoning.

\textbf{Third}, we leverage commercial MLLM APIs~\cite{reid2024gemini,2023gpt4V} to generate questions and answers by feeding the API with extracted evidence and carefully engineered prompts. We implement specific prompts for different types of evidence, including the task definition, requirements and expected output. \textit{For question generation}, the task definition is to let it generate two academic questions based on the given evidence and require the generation process to rely only on the provided content and eliminate other knowledge. Questions must be brief and objective and fully account for the provided evidence. \textit{For answer generation}, the model requires thinking step by step based on the question and evidence and generating two answers. The expected outputs include a long answer and a short answer, each with its own thought process. Short answers are used for training and evaluation of PDF-WuKong. Long answers and thought processes can be used to further optimize models to have document reasoning capabilities in the future. 

For detailed prompt engineering and the Query-Evidence-Answer (Q-E-A) triplet examples of each type, please refer to the Appendix~\ref{sec:appendix}.
Given the evidence and the prompt as the input, we use Gemini Pro~\cite{reid2024gemini} to generate questions and answers for the training set due to its free and rapid accessibility. GPT-4V~\cite{2023gpt4V} is used for the test set to ensure high evaluation quality. Using different APIs causes data bias but poses a
higher challenge to the generalizability of the model.

\begin{table}[h]
\centering
% \small
\renewcommand{\arraystretch}{1.2}
\caption{The statistics of QA pairs for various
types of evidence in English and Chinese. }
\begin{tabular}{>{\centering\arraybackslash}p{2cm}c|>{\centering\arraybackslash}p{2cm}|>{\centering\arraybackslash}p{2cm}}
\toprule
\multicolumn{2}{c|}{\textbf{Category}}& \multicolumn{1}{c|}{\textbf{Train (En/Zh)}} & \multicolumn{1}{c}{\textbf{Test (En/Zh)}} \\ 
\hline\hline
\addlinespace[2pt]
\multicolumn{1}{c|}{\multirow{2}{*}{\rotatebox{90}{\textit Single}}}    & Text-only    & 249k/12k  & 2939/296  \\
\multicolumn{1}{c|}{}  & Image-only  & 21K/40k  & 212/1018  \\
\addlinespace[2pt]
\hline
\addlinespace[2pt]
\multicolumn{1}{c|}{\multirow{3}{*}{\rotatebox{90}{\textit Multi}}}  & Image-text      & 250k/53k  & 2566/2150  \\
\multicolumn{1}{c|}{}     & Section & 499k/7k  &  255/394  \\
\multicolumn{1}{c|}{}   & Cross-paragraph   & 1.2k/0  & 118/0  \\
%\addlinespace[2pt]
\bottomrule
\end{tabular}
%\captionsetup{skip=10pt} 
\label{tab:data}
\end{table}

\setlength{\tabcolsep}{2.pt}
\begin{table}[t!]
% \small
\caption{Statistical analysis compared with other multi-page document datasets. \textbf{Page} and \textbf{Token} represent the average number of pages per document, and the average number of OCR tokens per document.} 
\begin{center}
\begin{tabular}{lllccc}
\toprule
\textbf{MPdataset}
& \textbf{Doc.} & \textbf{QA} & \textbf{Lang.} & \textbf{Page}  &   \textbf{Token}\\
\midrule
DUDE~\cite{van2023document}  & 5k & 41.5k & En & 6  &1831   \\
MP-Docvqa~\cite{TITO2023109834} & 6k & 46k & En & 8 &2026 \\
SlideVQA~\cite{SlideVQA2023} & 2.6k & 14.5k & En & 20 &1489\\
% GPT-4  & - & - & - & -  \\
\midrule
% \rowcolor[HTML]{F2F3F5}
$\text{PaperPDF (En)}$  & 60k & 900\revise{k}& En & 25  & 11371   \\ 
$\text{PaperPDF (Zh)}$  & 10k & 200k& Zh & 11 & 3413    \\ 
$\text{PaperPDF}$  & 70k & 1.1M & En+Zh & 23&  10234   \\ 
\bottomrule
\end{tabular}

\label{statictic_tab}
\end{center}
\end{table}

\textbf{Fourth}, to ensure data quality, we conduct data filtering for the Q-E-A triplets. We designed many effective rules to automatically eliminate low-quality samples and remove abnormal data. For example, too-short questions or too-long answers may confuse the model, and duplicate questions will reduce diversity. After performing automatic filtering for the training and testing data, we performed manual filtering and correction for the testing data to improve the quality further. \revise{This human verification step is crucial to mitigate potential biases inherent in model-generated content. By having human experts verify and correct the Q-E-A triplets, we ensure that the test set ground truth reflects accurate human reasoning rather than the stylistic patterns of a specific commercial API.} We checked if the questions made sense, if the answers were correct, if the Q-E-A triplets were related to each other, etc. We will present the statistical analysis and quality verification of the dataset in the next section.

\subsection{Dataset analysis}

In total, we obtained $1.1M$ bilingual training data and $10k$ testing data. The number of QA pairs for various types of evidence is shown in Tab.~\ref{tab:data}. To comprehensively illustrate the characteristics of the PaperPDF dataset, we conduct a detailed statistical analysis in Tab.~\ref{statictic_tab}. Compared to previous multi-page document datasets, PaperPDF not only includes a significantly larger number of documents and QA pairs but also features more pages and OCR tokens. 
More importantly, PaperPDF encompasses both Chinese and English documents with interleaved text and images.

\begin{figure}[ht]
\centering
\includegraphics[scale=0.4]{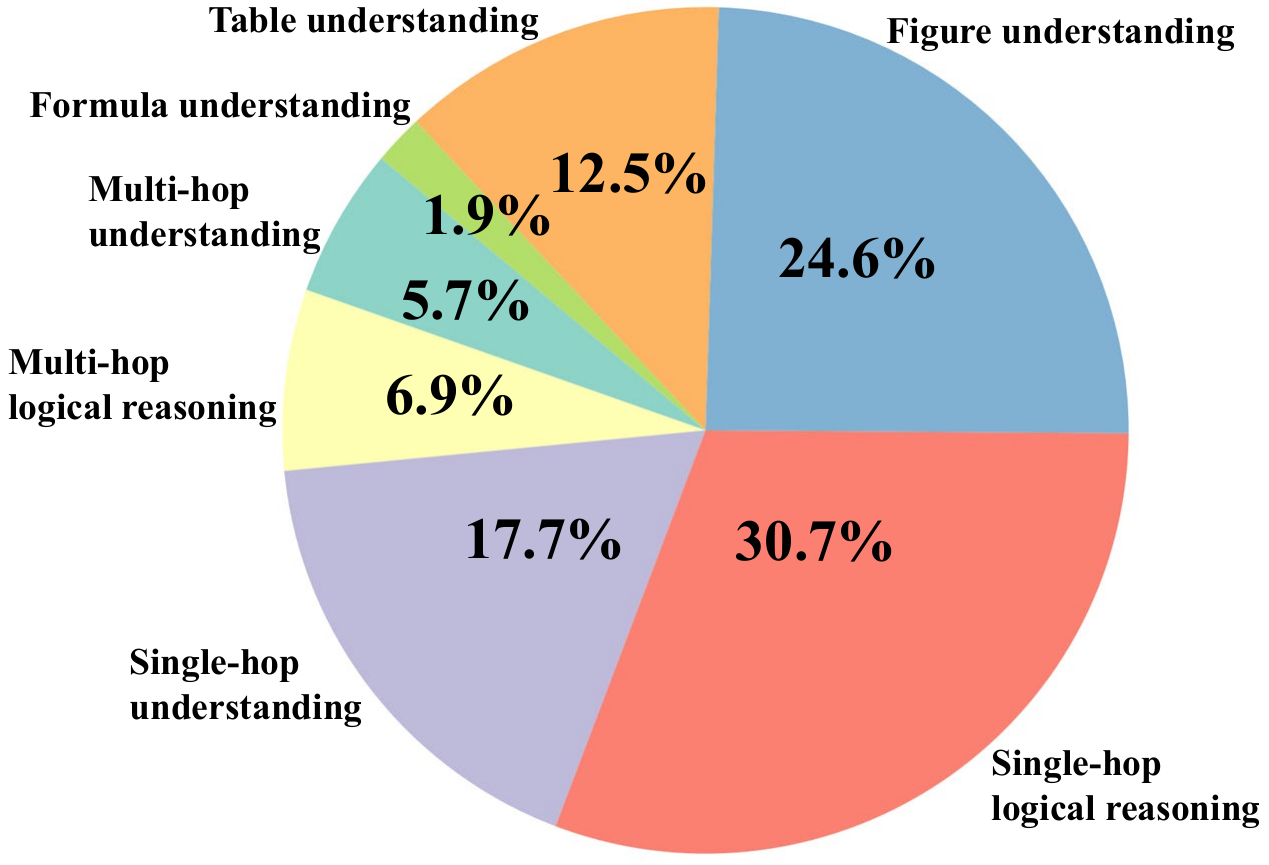}
\caption{Statistics of question types in PaperPDF dataset.}
\label{fig:data_sta}
\end{figure}

Fig.~\ref{fig:data_sta} shows the distribution of question types in our benchmark. We basically divide the types of questions in multimodal PDF documents into 7 categories, covering various questions that users may encounter. In order to further verify the correctness of the answers corresponding to each type of question, we manually evaluated the answers for each type in the test set. We first randomly selected 100 samples from each type and then asked two doctoral students with scientific research training to give an answer to each question. If the two people's answers disagree, they need to discuss with each other to determine the final answer. Finally, we counted the consistency between the answers from the test set and the answers given by humans, as shown in Tab.~\ref{data_evl}. The high agreement in the answers demonstrates the high quality of our benchmark.

Our dataset provides an important innovation driver and a comprehensive benchmark for the development of the multimodal long PDF document understanding task. Explicit constructing Q-E-A triplets aligns with PDF-WuKong's dual training objectives and ensures PDF-WuKong learns to distinguish fine-grained evidence patterns essential for long-document QA.

\begin{table}[t!]
% \small
\caption{Answers' consistency from the testing dataset and human evaluation for different types of questions.} 
\begin{center}
\begin{tabular}{lc}
\toprule
\textbf{Question Type} & \textbf{Consistency}  \\
\midrule
Single-hop understanding  & 95.2\%    \\
Single-hop logical reasoning & 93.0\%   \\
Multi-hop understanding  & 93.6\%    \\ 
Multi-hop logical reasoning  & 92.3\%     \\ 
Figure understanding  & 94.8\%  \\ 
Table understanding  & 97.3\%  \\ 
Formula understanding  & 96.1\%  \\ 
\bottomrule
\end{tabular}

\label{data_evl}
\end{center}
\end{table}

\section{Experiments}
\label{sec:exp}

\setlength{\tabcolsep}{2.5pt}
\begin{table*}[t!]
% \vspace{-4ex}
% \small
\caption{Performance comparison with open-source models for long PDF understanding on PaperPDF. * indicates that the RAG model has been trained on our dataset. 2p and 3p indicate whether VisRAG retrieves 2 or 3 pages of evidence. The best results are marked \textbf{bold} and the second results are \underline{underlined}.} 
\begin{center}
\begin{tabular}{llccccccccc}
\toprule
\multirow{2}{*}{\textbf{Model}}&  \multirow{2}{*}{\textbf{\#Param}} 
&  \multicolumn{4}{c}{\textbf{English}}&\multicolumn{4}{c}{\textbf{Chinese}}&\multirow{2}{*}{\textbf{Token}} \\
\cmidrule(lr){3-6} \cmidrule(lr){7-10}
&&\textbf{ANLS} & \textbf{F1} & \textbf{Rouge}  & \textbf{GPT-Acc}  &
 \textbf{ANLS} & \textbf{F1} & \textbf{Rouge}& \textbf{GPT-Acc} &   \\
\midrule
\multicolumn{11}{c}{\textit{Plain Text Solution}}\\ \midrule   % 建议居中
$\text{IXC2-VL~\cite{internlmxcomposer2}}$  
& 8B& 27.4 & 30.8 & 32.8 & 37.6 
& 21.1 & 28.5 & 28.7 & 30.8 & 4644  \\
$\text{IXC2-VL~\cite{internlmxcomposer2}+BGE-M3~\cite{chen2024bge}}$ 
& 8.5B& 32.4 & 34.0 & 32.4 &   48.4
& 23.3 & 29.8 & 32.3 & 38.1 & \underline{623} \\
%$\text{IXC2-4KHD~\cite{internlmxcomposer2_4khd}}^{\intercal}$ &8B&25.4&21.5&20.5&26.6&21.5&23.9&23.5&26.7&5010\\
%$\text{IXC2-4KHD-RAG~\cite{internlmxcomposer2_4khd}}^{\intercal}$ &8.5B&25.5&21.0&20.1&30.0&24.1&25.2&24.1&43.0&758\\
$\text{InternVL2~\cite{chen2024far}}$ 
&8B &19.5& 28.0&27.6&37.5
&24.4&28.2&27.7&30.8&4051\\
$\text{InternVL2~\cite{chen2024far}+BGE-M3~\cite{chen2024bge}}$ 
&8.5B&29.8&29.8&28.3&50.2&
24.6&27.8&26.8&43.0&\textbf{583}\\
$\text{InternVL2~\cite{chen2024far}+BGE-M3*~\cite{chen2024bge}}$ 
&8.5B&36.4&37.3&34.7&53.3&
28.4&34.2&34.5&46.3&849\\
\midrule
\multicolumn{11}{c}{\textit{Pure Vision Solution}} \\ \midrule
 $\text{Hi-VT5~\cite{TITO2023109834}}$  &0.3B & 13.5 & 3.1 & 3.7 & 15.2&-&-&-&-& 11589   \\
$\text{IXC2-VL~\cite{internlmxcomposer2}}$
&8B&27.1&24.8&25.1&20.5&15.9&19.5&22.2&27.4&4712
 \\
$\text{InternVL2~\cite{chen2024far}}$ 
&8B & 29.4 & 33.1 & 35.0  & 35.5&
20.3 & 33.9 & 37.8 & 37.7 & 5008  \\
$\text{Qwen2.5-VL~\cite{Qwen2.5-VL}}$ 
&7B & 35.5 & 34.8 & 32.6  & 38.3&
23.5 & 34.4 & 38.7 & 40.8 & 5878  \\
$\text{MiniCPM-V 2.6~\cite{yao2024minicpm} + VisRAG*~\cite{yu2024visrag}(2p)}$ 
&10B & 31.8 & 31.0 & 27.2  & 27.6&
20.5 & 27.0 & 30.1 & 35.9 & 1444  \\
$\text{MiniCPM-V 2.6~\cite{yao2024minicpm} + VisRAG*~\cite{yu2024visrag}(3p)}$ 
&10B & 32.4 & 32.5 & 28.5  & 28.4&
20.8 & 27.9 & 30.8 & 37.7 & 2199  \\
\midrule
\multicolumn{11}{c}{\textit{Parsed Interleaved Text and Images}}\\ \midrule
$\text{IXC2-VL~\cite{internlmxcomposer2}}$  
 &8B& 27.8& 31.2& 32.6 & 37.7 
 &22.5 & 29.5 & 29.2 & 31.4 & 6217  \\
% &$\t
$\text{InternVL2~\cite{chen2024far}}$
&8B& \underline{33.4} & \underline{36.2}  & \underline{36.6}  & \underline{54.3}
&\underline{28.5} & \underline{40.6} & \underline{42.0}  & \underline{54.7} & 6220 \\
\rowcolor[HTML]{F2F3F5}
$\text{PDF-WuKong (ours)}$  
& 8.5B& \textbf{41.9} & \textbf{43.5} & \textbf{40.9} &
\textbf{77.5}&
\textbf{{40.9}} & \textbf{{47.8}} & \textbf{{48.6}} & \textbf{57.8} &2107 \\ 
\bottomrule
\end{tabular}

\label{main_tab}
\end{center}
\end{table*}

\subsection{Implementation Details}

% The LLM and the vision encoder are initialized from IXC2-VL-4KHD~\cite{internlmxcomposer2_4khd} and the maximum number of dynamic tiles is set as 16. The text encoder is initialized from BGE-M3~\cite{chen2024bge}. We train the model by leveraging several document datasets including PaperPDF, DocVQA~\cite{Mathew_2021_WACV}, ChartQA~\cite{masry2022chartqa}, InfoVQA~\cite{Mathew_2022_WACV}, MPDocVQA~\cite{TITO2023109834}, and DUDE~\cite{van2023document}. Before both training and testing, PaperPDF is parsed using Grobid~\cite{GROBID} and MinerU~\cite{wang2024mineruopensourcesolutionprecise}, while the other datasets are processed following their default instructions. The training is conducted for one epoch using 128 Ascend 910B NPUs with a learning rate of 4e-5. We set the top 5 sampling results as input to the LLM. 

\revise{The LLM and the vision encoder are initialized from IXC2-VL-4KHD~\cite{internlmxcomposer2_4khd} and the maximum number of dynamic tiles is set as 16. The text encoder is initialized from BGE-M3~\cite{chen2024bge}. We train the model by leveraging several document datasets including PaperPDF, DocVQA~\cite{Mathew_2021_WACV}, ChartQA~\cite{masry2022chartqa}, InfoVQA~\cite{Mathew_2022_WACV}, MPDocVQA~\cite{TITO2023109834}, and DUDE~\cite{van2023document}. Before both training and testing, PaperPDF is parsed using Grobid~\cite{GROBID} and MinerU~\cite{wang2024mineruopensourcesolutionprecise} to support sparse sampling. In contrast, for the other datasets, we utilize the complete full-page images paired with their corresponding OCR text as global input blocks, bypassing the fine-grained parsing step.} The training is conducted for one epoch using 128 Ascend 910B NPUs with a learning rate of 4e-5. We set the top 5 sampling results as input to the LLM. 

For the model evaluations on the PaperPDF dataset, we use three objective metrics ANLS, F1, and Rouge to report the quantitative results on the full test set. To evaluate the semantic correctness of the answer, we introduce GPT-Acc to determine whether the output is correct. Considering the expensive cost of GPT-4 evaluation, we randomly selected two subsets with 50 English PDFs and 30 Chinese PDFs for GPT-Acc calculation. They contain 488 and 317 QA samples, respectively.

\begin{figure*}[t!]
\centering
\includegraphics[width=1.0\textwidth]{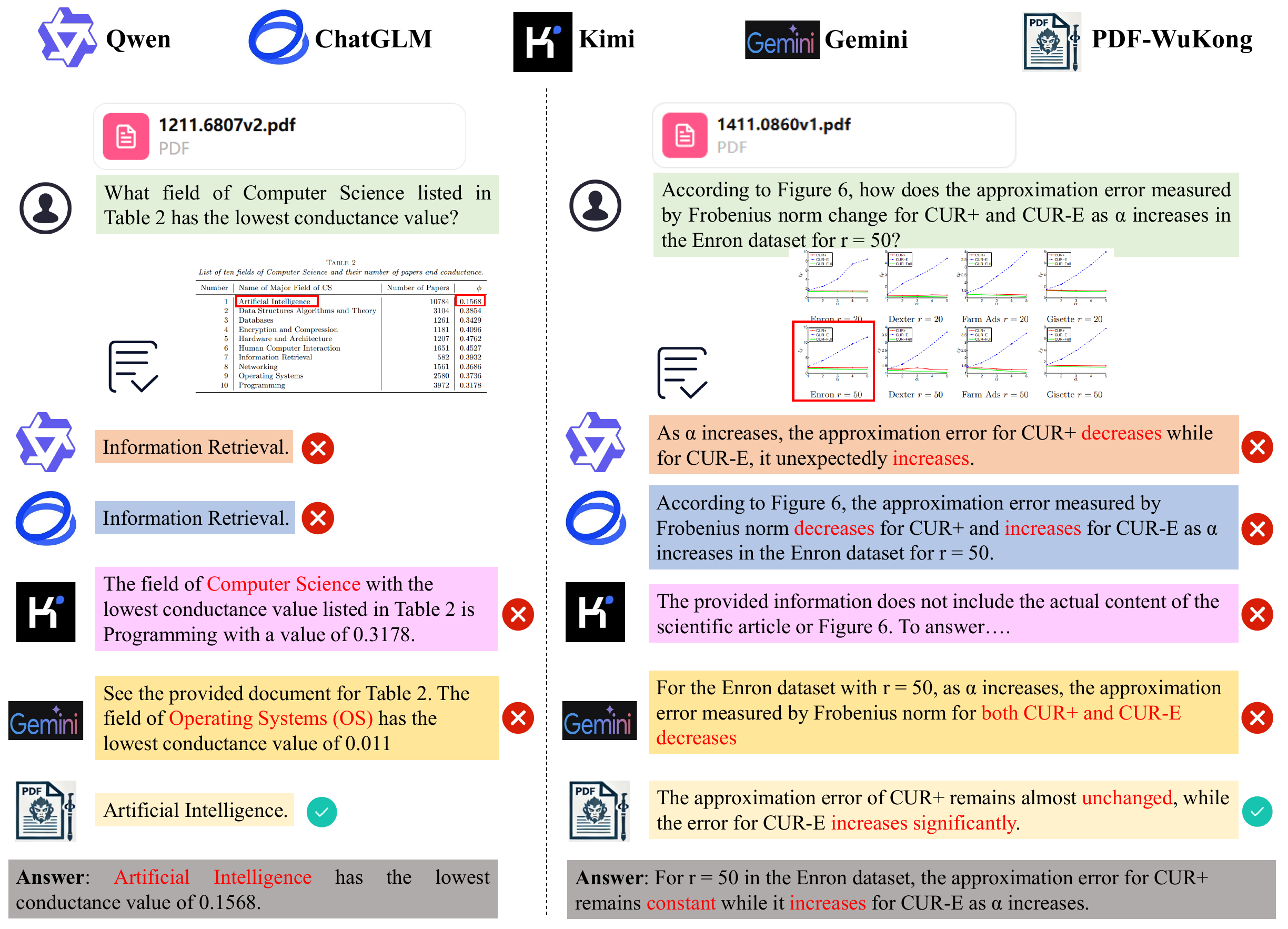}
\caption{Qualitative comparison between PDF-WuKong with other proprietary products. The red box indicates the evidence that the correct answer depends on. Our model can find the text chunks, figures or tables contaning the evidence and output correct answers.}
\label{fig:example}
\end{figure*}

\subsection{Long PDF Understanding}

To assess the effectiveness of our model in understanding long PDF documents, we conduct comprehensive experiments comparing it with both open-source models and commercial products on the PaperPDF dataset. 

Due to the limited capabilities of traditional document understanding models  ~\cite{TITO2023109834, blau2024gram, huang2022layoutlmv3, beltagy2020longformer, zaheer2020big} and the huge resource costs in advanced MLLMs ~\cite{internlmxcomposer2, internlmxcomposer2_4khd, chen2024far}, achieving a deep understanding of lengthy PDF documents remains a highly challenging task. To handle this task, we explore \textbf{three approaches} to input the PDF documents into these MLLMs: \textbf{pure text content, page images, and parsed interleaved text-image content}. For some baselines with plain text modality input, we also report the results based on Text RAG. For the pure vision solutions, we compared some document understanding models that can accept multi-page image input. \revise{To ensure a fair comparison regarding data availability, we not only compared against off-the-shelf RAG models but also fine-tuned the retrievers of the RAG baselines (BGE-M3 ~\cite{chen2024bge}, VisRAG~\cite{yu2024visrag}) on the PaperPDF dataset.}

\setlength{\tabcolsep}{2.pt}
\begin{table}[t!]
% \small
\caption{Comparison with commercial products (tested in Sep 2024) for long PDF understanding. The results are tested on a subset of 50 English PDFs. Note: These are the products rather than the models themselves.}
% \vspace{-4ex}
\begin{center}
\begin{tabular}{llcccc}
\toprule
\textbf{Model}
& \textbf{ANLS} & \textbf{F1} & \textbf{Rouge}  & \textbf{GPT-Acc}  \\
  \midrule
Gemini 1.5 pro~\cite{gemini}  & 26.6 & 29.0 & 29.8& 67.9   \\
 Kimi~\cite{kimi4} & 28.5 & 33.6 & 31.1 & 74.7  \\
 ChatGLM~\cite{chatglm}  & 31.2 & 35.4 & 32.0& 73.5 \\
Qwen~\cite{tongyi} &\underline{36.0} & \underline{40.3} & \underline{35.5}& \textbf{78.1}  \\
% GPT-4  & - & - & - & -  \\
\rowcolor[HTML]{F2F3F5}
 $\text{PDF-WuKong (ours)}$  & \textbf{{41.8}} & \textbf{{43.2}} & \textbf{{40.7}}& \underline{77.5}   \\ 
\bottomrule
\end{tabular}
 
\label{closed_tab}
\end{center}
\end{table}

\begin{figure*}[htbp]
\centering
\includegraphics[width=0.95\textwidth]{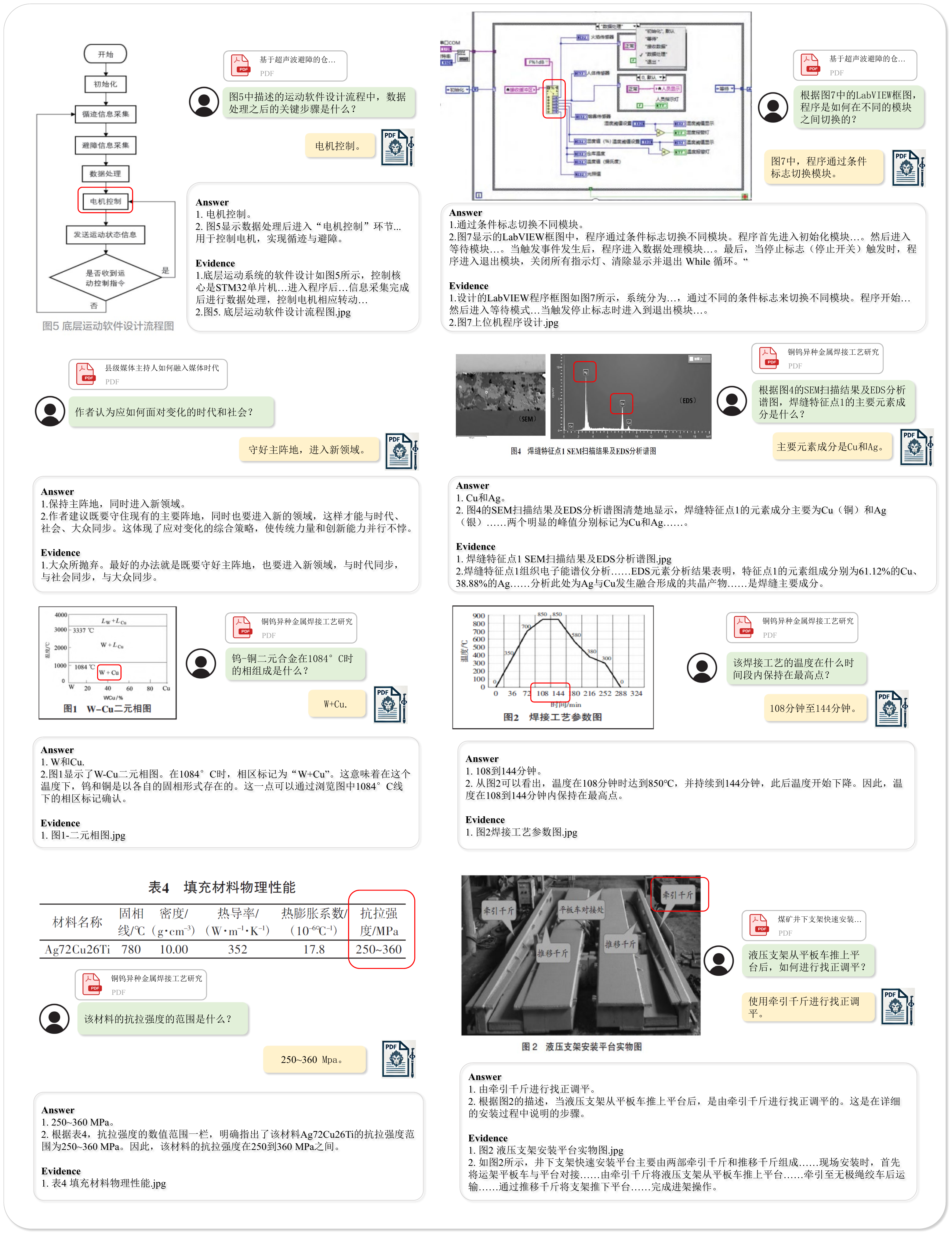}
\caption{Examples of PDF-WuKong on Chinese documents. The red box indicates the evidence that the correct answer depends on.}
\label{fig:ch_visualization}
\end{figure*}

The experimental results for the open-source models are reported in Tab.~\ref{main_tab}. \revise{The performance gap between PDF-WuKong and these fine-tuned baselines cannot be attributed solely to data exposure. Instead, it validates the architectural superiority of our end-to-end sparse sampling mechanism over disjoint retrieval-augmented pipelines.} Generally, both document understanding models and RAG models oriented towards \textbf{a single modality} are inferior to our paradigm. 
Moreover, we compare PDF-WuKong with some closed-source commercial products in Tab.~\ref{closed_tab}, also showing our superiority. The qualitative comparisons between PDF-WuKong and these proprietary products on the long PDF understanding task are presented in Fig.~\ref{fig:example} (English) and Fig.~\ref{fig:ch_visualization} (Chinese).

% Several key conclusions can be drawn from the results.  Firstly, inputting parsed interleaved text-image content generally outperforms multiple page images. 
% The main reason is the limited input resolution and number of tokens that the model can accept. 
% Secondly, when handling tokens of similar scale, inputting parsed interleaved text-image content yields better performance compared to pure text content. It is obvious that diagram information in the PDF plays a crucial role in document comprehension. This also indicates that PaperPDF places rigorous requirements on the visual information within documents. Additionally, we observe that for some baseline models, the approach of parsing interleaved text-image content as input outperforms other single-modal inputs. 
% Finally, benefiting from the inclusion of the spare sampler, our proposed PDF-WuKong model not only surpasses the existing state-of-the-art open-source model InternVL2 by approximately 7\% on both the Chinese and English subsets but also demonstrates competitive performance comparable to proprietary products. 
% Moreover, due to the integration of the sparse sampler, PDF-WuKong maintains efficiency in inference token cost.

Several key conclusions can be drawn from the results.

\noindent\textbf{1. Interleaved text-image content outperforms plain text input:} When handling tokens of similar scale, inputting parsed interleaved text-image content yields better performance compared to pure text content. It is obvious that diagram information in the PDF plays a crucial role in document comprehension. This also indicates that PaperPDF places rigorous requirements on the visual information within documents.

\noindent\textbf{2. Interleaved text-image content outperforms pure vision input:} When compared to pure visual input, interleaved text-image content provides superior performance. The main reason is the limited input resolution and number of tokens that the model can handle. 

\noindent\textbf{3. Sampling enhances understanding:} By introducing the single-modality RAG for MLLM to locate and sample evidence, their ability to understand long documents can be improved, which preliminarily verifies the feasibility of our motivation.

\noindent\textbf{4. Limitations of single-modal sampling:} Text-only RAG cannot focus on visual information, while visual-only RAG retrieves page images and cannot extract more fine-grained evidence. The full-page dense text images bring greater challenges to the image encoder.

\noindent\revise{\textbf{5. Superior performance on long PDF understanding:} Benefiting from the inclusion of the multi-modal spare sampler, our proposed PDF-WuKong not only surpasses InternVL2 by approximately 7\% on both the Chinese and English subsets but also demonstrates competitive performance compared with proprietary products.} Moreover, due to the integration of the end-to-end sparse sampler, PDF-WuKong maintains efficiency during training and inference with fewer tokens.

\setlength{\tabcolsep}{3pt}
\begin{table}[ht]
% \small
\caption{Performance comparison with other DocVLMs for PDF multimodal understanding on the Single-Evidence Subset. $\dagger$ denotes the page image input, aligning with other models in the table; * indicates that our model utilizes parsed content as input.} 
% \vspace{-4ex}
\begin{center}
\begin{tabular}{lrcccccc}
\toprule
 % \multirow{2}{*}{\textbf{Model}}& \multirow{2}{*}{\textbf{\#param}} & \multicolumn{3}{c}{\textbf{Single-Evidence Subset}}\\
  % & & \textbf{ANLS} & \textbf{F1} & \textbf{ROUGE}   \\ 
 \textbf{Model}& \textbf{\# param} & \textbf{ANLS} & \textbf{F1} & \textbf{ROUGE}\\
% \cmidrule{1-6}
\midrule
 $\text{Qwen-VL~\cite{Qwen-VL}}$ & 9.6B & 26.4 & 19.6 & 18.3 &  \\
 $\text{Monkey~\cite{li2023monkey}}$  & 9.8B& 30.0 & 24.4 & 22.3 &  \\
 $\text{mPLUG-Owl2~\cite{hu2024mplugdocowl2}}$ & 8.2B & 19.5 & 20.3 & 22.7 &  \\
 $\text{Emu2-Chat~\cite{sun2024generative}}$  & 37B & 26.0 & 24.4 & 23.4 &  \\

$\text{MiniCPM-2.5~\cite{hu2024minicpm}}$ & 8.5B & 31.8 & 28.2 & 24.8 &  \\
%& DeepSeek-vl-7b & 7B& - & - & -  \\
 $\text{IXC2-VL~\cite{internlmxcomposer2}}$  & 8B& 23.4 & 20.8 & 21.3 &  \\
 $\text{IXC2-4KHD~\cite{internlmxcomposer2_4khd}}$  & 8B& 24.5 & 20.0 & 18.0   \\
 $\text{CogVLM2~\cite{hong2024cogvlm2}}$ & 17B & 24.8 & 27.4 & 26.3 &  \\
 \midrule
 \rowcolor[HTML]{F2F3F5}
$\text{PDF-WuKong (ours)} ^ \dagger$  & 8.5B & \underline{36.6} & \underline{35.2} & \underline{31.7} & \\
\rowcolor[HTML]{F2F3F5}
$\text{PDF-WuKong (ours)} ^ *$  & 8.5B & \textbf{41.5} & \textbf{42.8} & \textbf{39.8} & \\

\bottomrule
% \footnotesize{($\dagger$) indicates the use of a special input approach. For details, refer to Table \ref{tab:input_symbols}.}
\end{tabular}

\label{single}
\end{center}
\end{table}

The main reason for not using our data to train other MLLMs is that they cannot receive lengthy multimodal documents as input. Most models only support single-page input for training and few models can handle up to 8 or 20 pages. They are unable to be trained on our data.
To compare with more open-source document MLLMs, we construct a subset of the PaperPDF benchmark, which only contains test samples with single evidence. \revise{This subset serves as a critical benchmark for fair comparison. It allows us to compare the fundamental multimodal understanding capabilities of our PDF-WuKong against other models under the same single-page input setting.}
Therefore, we feed all models with only one page containing the evidence as their input. The pages are input in the form of images. 
Our PDF-WuKong can accept input in two formats. One is the page image and another is the parsed page. 
As shown in Tab.~\ref{single}, our model's capability on this subset is significantly better than other document models. Moreover, introducing a parsed-input paradigm significantly reduces the difficulty of understanding text-rich documents, yielding approximately a 7\% performance improvement on the Single-Evidence subset. This improvement can be primarily attributed to two factors. First, the parsed input allows the model to sample more fine-grained evidence for answer generation. Second, compared to raw full-page images, the parsed image data substantially relaxes the model’s dependency on high-resolution input, thereby enhancing both efficiency and robustness.

\subsection{Document-oriented VQA}
To validate the strong capability of our model in other document understanding scenarios, we conduct experiments on several public benchmarks. \revise{It is worth noting that while our primary pipeline utilizes parsed PDF content, PDF-WuKong is natively compatible with non-PDF inputs. In the following experiments, we feed the model with page images rather than parsed PDF blocks. PDF-WuKong treats each page as a visual block and can directly retrieve the pages containing the evidence.}

First, we evaluate the performance of PDF-WuKong on single-page document datasets~\cite{Mathew_2021_WACV, masry2022chartqa, Mathew_2022_WACV}. As shown in Tab.~\ref{docvqa}, our model achieves comparable performance on single-page document understanding. This demonstrates that PDF-WuKong can effectively handle various types of documents and questions, showcasing its versatility in document-oriented visual question-answering tasks.

\setlength{\tabcolsep}{3pt}
\begin{table}[t]
% \small
\caption{Performance comparison with other DocVLMs on single-page document-oriented VQA benchmarks. } 
% \vspace{-4ex}
\begin{center}
\begin{tabular}{lccc}
\toprule
 \textbf{Model}  & \textbf{DocVQA} & \textbf{ChartQA} & \textbf{InfoVQA} \\
%\cline{5-6} \cline{7-8}
%\cline{2-3} \cline{4-4} \cline{5-5} \cline{6-6} \cline{7-8} 
%& \multicolumn{1}{c}{ANLS} & \multicolumn{1}{c}{Acc} & \multicolumn{1}{c}{Acc}  & \multicolumn{1}{c}{AP} & %\multicolumn{1}{c}{GPT-acc}& \multicolumn{1}{c}{ANLS} & \multicolumn{1}{X}{ECE} \\
\midrule
$\text{Qwen-VL~\cite{Qwen-VL}}$  & 65.1  & 65.7 & 35.4  \\
$\text{Monkey~\cite{li2023monkey}}$  & 66.5  & 65.1 & 36.1  \\
$\text{Text-Monkey~\cite{liu2024textmonkey}}$  & 73.0  & 66.9 & 28.6  \\
$\text{MiniCPM-V-2.5~\cite{hu2024minicpm}}$ & 84.8 & - & - \\
$\text{Vary-base~\cite{wei2025vary}}$ & 76.3 & 66.1 & - \\
$\text{TextHawk~\cite{yu2024texthawk}}$ &76.4&66.6&50.6\\
$\text{IXC2-4KHD-16~\cite{internlmxcomposer2_4khd}}$  & 84.9  & \textbf{80.1} & 60.8  \\
$\text{DocOwl 2~\cite{hu2024mplugdocowl2}}$  & 80.7  & 70.0& 46.4 \\
$\text{CREAM~\cite{zhang2024cream}}$ &79.4&-&53.6\\
% \hline
\midrule
\rowcolor[HTML]{F2F3F5}
$\text{PDF-WuKong (ours)}$  & \textbf{85.1}  & 80.0 & \textbf{61.3}  \\ 
\bottomrule
\end{tabular}

\label{docvqa}
\end{center}

\end{table}

\setlength{\tabcolsep}{3pt}
\begin{table}[t]
% \small
\caption{Performance comparison with other DocVLMs for multi-page document understanding.}
% \vspace{-4ex}
\begin{center}
\begin{tabular}{lcc}
\toprule
\textbf{Model}  &  \textbf{MP-DocVQA} & \textbf{DUDE} \\
%\cline{5-6} \cline{7-8}
%\cline{2-3} \cline{4-4} \cline{5-5} \cline{6-6} \cline{7-8} 
%& \multicolumn{1}{c}{ANLS} & \multicolumn{1}{c}{Acc} & \multicolumn{1}{c}{Acc}  & \multicolumn{1}{c}{AP} & %\multicolumn{1}{c}{GPT-acc}& \multicolumn{1}{c}{ANLS} & \multicolumn{1}{X}{ECE} \\
\midrule
\text{LayoutLMv3~\cite{huang2022layoutlmv3}} & 55.1 & 20.3 \\
$\text{Longformer~\cite{beltagy2020longformer}}$ & 55.1 & 27.1 \\
$\text{BigBird~\cite{zaheer2020big}}$ & 58.5 & 26.3 \\
$\text{Hi-VT5~\cite{TITO2023109834}}$ & 61.8 & 35.7  \\
$\text{DocFormerv2~\cite{appalaraju2024docformerv2}}$ &76.4 &48.4 \\
$\text{GRAM~\cite{blau2024gram}}$ &\textbf{83.0} &53.4 \\
% Gemini pro &-&- \\
$\text{GPT-4V (2024-06)~\cite{2023gpt4V}}$  &-& \underline{53.9} \\
$\text{Idefics3-8B~\cite{laurenccon2024obelics}}$ & 67.2 & 38.7 \\
$\text{DocOwl2~\cite{hu2024mplugdocowl2}}$ &69.4 & 46.7 \\
$\text{CREAM~\cite{zhang2024cream}}$ &65.3&52.5\\
SV-RAG~\cite{chen2025svrag} &71.0 &45.0 \\
\midrule
\rowcolor[HTML]{F2F3F5}
% $\text{PDFMaster}^\intercal$  & 68.1 &53.0\\ 
% \rowcolor[HTML]{F2F3F5}
% $\text{PDFMaster}^\dagger$  & 70.3 &53.6\\ 
% \rowcolor[HTML]{F2F3F5}
$\text{PDF-WuKong (ours)}$  & \underline{76.9} &\textbf{56.1}\\ 
\bottomrule
\end{tabular}

\label{multipage}
\end{center}
\end{table}

In addition, we assess the performance of specialized models and MLLMs on two existing multi-page document QA datasets. The results shown in Tab.~\ref{multipage} indicate that our model's performance in multi-page document scenarios is comparable to these specialized models and far surpasses the latest document MLLM DocOwl2~\cite{hu2024mplugdocowl2}. Notably, on complex multi-page document datasets like DUDE~\cite{van2023document}, PDF-WuKong outperforms GPT-4V~\cite{2023gpt4V}. This improvement is attributed to the sparse sampler, which extracts useful information from multi-page documents, enabling the model to focus on relevant content. 
% Besides, SV-RAG~\cite{chen2025svrag} performs much worse than PDF-WuKong on these important benchmarks. It treats each page as a separate image and uses pure visual retrieval to obtain page-level evidence. This further shows that the pure visual solution is inferior to ours based on multimodal retrieval for the parsed documents.

Furthermore, we conduct zero-shot evaluations on a long multimodal document understanding benchmark MM-NIAH~\cite{wang2024needle}. As shown in Tab.~\ref{niah}, our model uses the fewest parameters yet achieves the second-best performance. Although InternVL-V1.5-RAG surpasses PDF-WuKong by 2.8\%, it utilizes 36.5B more parameters than our model. Moreover, as the context length of the multimodal documents increases, the performance of our model remains stable, while that of other models significantly decreases. At a context length of 64K, PDF-WuKong even achieves the best performance, demonstrating its robustness in handling long-context multimodal inputs.

\setlength{\tabcolsep}{1pt}
\begin{table}[t]
% \small
\caption{Performance comparison with other DocVLMs on MM-NIAH. The evaluation approach aligns with the benchmark.}
% \vspace{-4ex}
\begin{center}
\begin{tabular}{lccccccc}
\toprule
\textbf{Model}  & \textbf{\#param}&  \textbf{Overall} & \textbf{1K} & \textbf{4K}& \textbf{16K}&\textbf{64K}\\
%\cline{5-6} \cline{7-8}
%\cline{2-3} \cline{4-4} \cline{5-5} \cline{6-6} \cline{7-8} 
%& \multicolumn{1}{c}{ANLS} & \multicolumn{1}{c}{Acc} & \multicolumn{1}{c}{Acc}  & \multicolumn{1}{c}{AP} & %\multicolumn{1}{c}{GPT-acc}& \multicolumn{1}{c}{ANLS} & \multicolumn{1}{X}{ECE} \\
\midrule
Emu2-Chat~\cite{sun2024generative} & 37B & 8.8 & 38.9 &18.2 &0.0&0.0 \\
% VILA1.0-13b-llava\cite{lin2024vila} & 13B & 15.7 & 41.9 & 33.2 & 8.6 & 0.1 \\
VILA1.0-13b~\cite{lin2024vila} & 13B & 15.7 & 41.9 & 33.2 & 8.6 & 0.1 \\
% llava-v1.6-vicuna-13b\cite{liu2023llava} & 13B &16.9 &43.7&34.9&13.6&0.0  \\
llava-v1.6-13b~\cite{liu2024llavanext} & 13B &16.9 &43.7&34.9&13.6&0.0  \\
llava-v1.6-34b~\cite{liu2024llavanext} & 34B &20.6 &\underline{57.4}&\underline{45.1}&8.2&0.0 \\
InternVL1.5~\cite{chen2024far} & 26B &41.1&\textbf{59.5}&\textbf{50.1}&41.9&16.6  \\
InternVL1.5-RAG~\cite{chen2024far} & 45B &\textbf{46.1}&\textbf{59.5}&\textbf{50.1}&\textbf{44.9}&39.3 \\
% Gemini-1.5  &54.9& 64.7 &56.8&53.7&49.8 \\
\midrule
\rowcolor[HTML]{F2F3F5}
PDF-WuKong (ours) & 8.5B &\underline{43.3} & 53.0 & 43.9 & \underline{43.0} & \textbf{42.1} \\ 
\bottomrule
\end{tabular}
 
\label{niah}
\end{center}
\end{table}

\subsection{\revise{Fine-grained Evaluation on Long PDF Understanding}}

\revise{To comprehensively assess the capabilities of PDF-WuKong in long-document processing, we conducted a fine-grained evaluation on the PaperPDF benchmark, as presented in Table~\ref{evaluation_on_diff_question}. We can  observe that the model exhibits superior performance in figure and table understanding tasks compared to other categories questions, validating that the sparse sampling mechanism effectively captures critical visual evidence within long documents. However, formula understanding remains weaker, which may be attributed to the loss of 2D structural semantics during text linearization and the scarcity of relevant training data. Additionally, the model performs slightly better on Multi-hop tasks, a trend suggesting that the model potentially benefits from utilizing global redundant context, unlike single-hop tasks that focus on locating isolated details.}

% \setlength{\tabcolsep}{3pt}
% \begin{table}[ht]
% % \small
% \caption{\revise{Performance of PDF-WuKong across different question types.}} 
% \label{evaluation_on_diff_question}
% % \vspace{-4ex}
% \begin{center}
% \begin{tabular}{lccc}
% \toprule
% \revise{{\textbf{Question Type}}}  &\revise{\textbf{ANLS}} & \revise{\textbf{F1}} & \revise{\textbf{ROUGE}}   \\
% %\cline{5-6} \cline{7-8}

% \midrule
% \revise{Table Understanding} & \revise{44.15} & \revise{46.32} & \revise{45.09}  \\
% \revise{Figure Understanding} & \revise{46.05} & \revise{48.38} & \revise{46.46}  \\
% \revise{Formula Understanding} & \revise{39.67} & \revise{39.81} & \revise{38.08}  \\
% \revise{Multi-hop Understanding} & \revise{44.07} & \revise{45.30} & \revise{45.01}  \\
% \revise{Multi-hop Logical Reasoning} & \revise{41.13} & \revise{42.40} & \revise{38.92}  \\
% \revise{Single-hop Understanding} & \revise{42.32} & \revise{43.33} & \revise{40.30}  \\
% \revise{Single-hop Logical Reasoning} & \revise{39.05} & \revise{38.41} & \revise{35.55}  \\
% \revise{Average} & \revise{42.59} & \revise{43.63} & \revise{40.19} \\
% %42.59 43.63 40.19
% \bottomrule
% \end{tabular}

% \end{center}
% \end{table}

\setlength{\tabcolsep}{3pt}
\begin{table}[t]
% \small
\caption{\revise{Performance of PDF-WuKong across different question types. TU, FU, FoU, MuU, MuL, SiU and SiL represent Table Understanding, Figure Understanding, Formula Understanding, Multi-hop Understanding, Multi-hop Logical Reasoning, Single-hop Understanding and Single-hop Logical Reasoning, respectively.}} 
\label{evaluation_on_diff_question}
% \vspace{-4ex}
\begin{center}
\begin{tabular}{lcccccc}
\toprule
\revise{{\textbf{Question}}}  &\multicolumn{3}{c}{\revise{PDF-WuKong}} &\multicolumn{3}{c}{\revise{InternVL2}}  \\
\cmidrule(lr){2-4}
\cmidrule(lr){5-7}
\revise{{\textbf{Type}}}  &\revise{\textbf{ANLS}} & \revise{\textbf{F1}} & \revise{\textbf{ROUGE}} &\revise{\textbf{ANLS}} & \revise{\textbf{F1}} & \revise{\textbf{ROUGE}}  \\
%\cline{5-6} \cline{7-8}

\midrule
\revise{TU} & \revise{44.15} & \revise{46.32} & \revise{45.09} & \revise{40.65} & \revise{45.44} & \revise{45.71}  \\
\revise{FU} & \revise{46.05} & \revise{48.38} & \revise{46.46} & \revise{40.82} & \revise{44.68} & \revise{45.03} \\
\revise{FoU} & \revise{39.67} & \revise{39.81} & \revise{38.08} & \revise{26.67} & \revise{28.59} & \revise{30.72}  \\
\revise{MuU} & \revise{44.07} & \revise{45.30} & \revise{45.01} & \revise{35.83} & \revise{37.07} & \revise{39.13} \\
\revise{MuL} & \revise{41.13} & \revise{42.40} & \revise{38.92}  & \revise{32.13} & \revise{33.37} & \revise{34.26} \\
\revise{SiU} & \revise{42.32} & \revise{43.33} & \revise{40.30}   & \revise{28.69} & \revise{31.11} & \revise{30.90} \\
\revise{SiL} & \revise{39.05} & \revise{38.41} & \revise{35.55}  & \revise{27.04} & \revise{28.16} & \revise{28.46} \\
%42.59 43.63 40.19
\bottomrule
\end{tabular}

\end{center}
\end{table}

\subsection{Ablation Study}

To comprehensively evaluate the effectiveness of our contributions, we conduct ablation studies focusing on the impact of the sparse sampler, the dataset, the document length, and sampling strategies. These experiments are based on the English subset of our PaperPDF for training and evaluation to avoid interference from other factors.

\noindent \textbf{Sparse sampler}

To assess the effectiveness of the sparse sampler, we compared models trained with and without it. Without the sparse sampler, the MLLM struggled to process long documents with interleaved text and images, resulting in poor performance due to the large amount of irrelevant information. Introducing the sparse sampler significantly improved the model's accuracy, as evidenced in Tab.~\ref{sampler}, by efficiently selecting the most relevant content for each query. Furthermore, end-to-end joint training of the sparse sampler and the MLLM led to additional performance gains compared to training them separately. 
This indicates that our end-to-end optimization of multimodal representation and question answering can further promote the document understanding ability of MLLM.

\setlength{\tabcolsep}{3pt}
\begin{table}[ht]
% \small
\caption{Ablation study on the impact of sparse sampler} 
% \vspace{-4ex}
\begin{center}
\begin{tabular}{cc|ccc}
\toprule
\textbf{Sparse Sampler} & \textbf{End-to-End}  & \textbf{ANLS} & \textbf{F1} & \textbf{ROUGE}   \\
%\cline{5-6} \cline{7-8}
%\cline{2-3} \cline{4-4} \cline{5-5} \cline{6-6} \cline{7-8} 
%& \multicolumn{1}{c}{ANLS} & \multicolumn{1}{c}{Acc} & \multicolumn{1}{c}{Acc}  & \multicolumn{1}{c}{AP} & %\multicolumn{1}{c}{GPT-acc}& \multicolumn{1}{c}{ANLS} & \multicolumn{1}{X}{ECE} \\
\midrule
\ding{55} & \ding{55} & 11.1 & 5.1 & 5.0  \\
\ding{51} & \ding{55} & 40.3 & 42.3 & 39.8  \\
\rowcolor[HTML]{F2F3F5}
\ding{51} & \ding{51} & \textbf{42.6} & \textbf{43.6} & \textbf{40.2}  \\
% GPT-4  & - & - & - & -  \\
\bottomrule
\end{tabular}

\label{sampler}
\end{center}
\end{table}

\noindent \textbf{Dataset}

We retrain PDF-WuKong on various dataset settings of the English subset of our PaperPDF, and the results are shown in Tab.~\ref{dataset}. Increasing the amount of training data leads to consistent improvements, proving that our dataset is of high quality and follows scaling laws.
Besides, we verify the effectiveness of our two data construction methods. Under the same data scale, multi-evidence data can enhance the model's complex reasoning ability.

\setlength{\tabcolsep}{3pt}
\begin{table}[t]
% \small
\caption{Ablation study on dataset setting} 
% \vspace{-4ex}
\begin{center}
\begin{tabular}{ccccc}
\toprule
\textbf{Dataset}  & \textbf{ANLS} & \textbf{F1} & \textbf{ROUGE}   \\
%\cline{5-6} \cline{7-8}

\midrule
100 k & 38.7 & 40.1 & 37.5  \\
500 k & 41.6 & 43.5 & 40.8  \\
1 M  & 42.6 & 43.6 & 40.2  \\
\midrule
Single (200k) & 39.2 & 40.4 & 38.1  \\
Multi (100k) + Single (100k) & 40.0 & 41.6 & 38.9  \\

\bottomrule
\end{tabular}

\label{dataset}
\end{center}
\end{table}

\noindent \textbf{Document length}

\begin{figure}[t]
\centering
\includegraphics[scale=0.62]{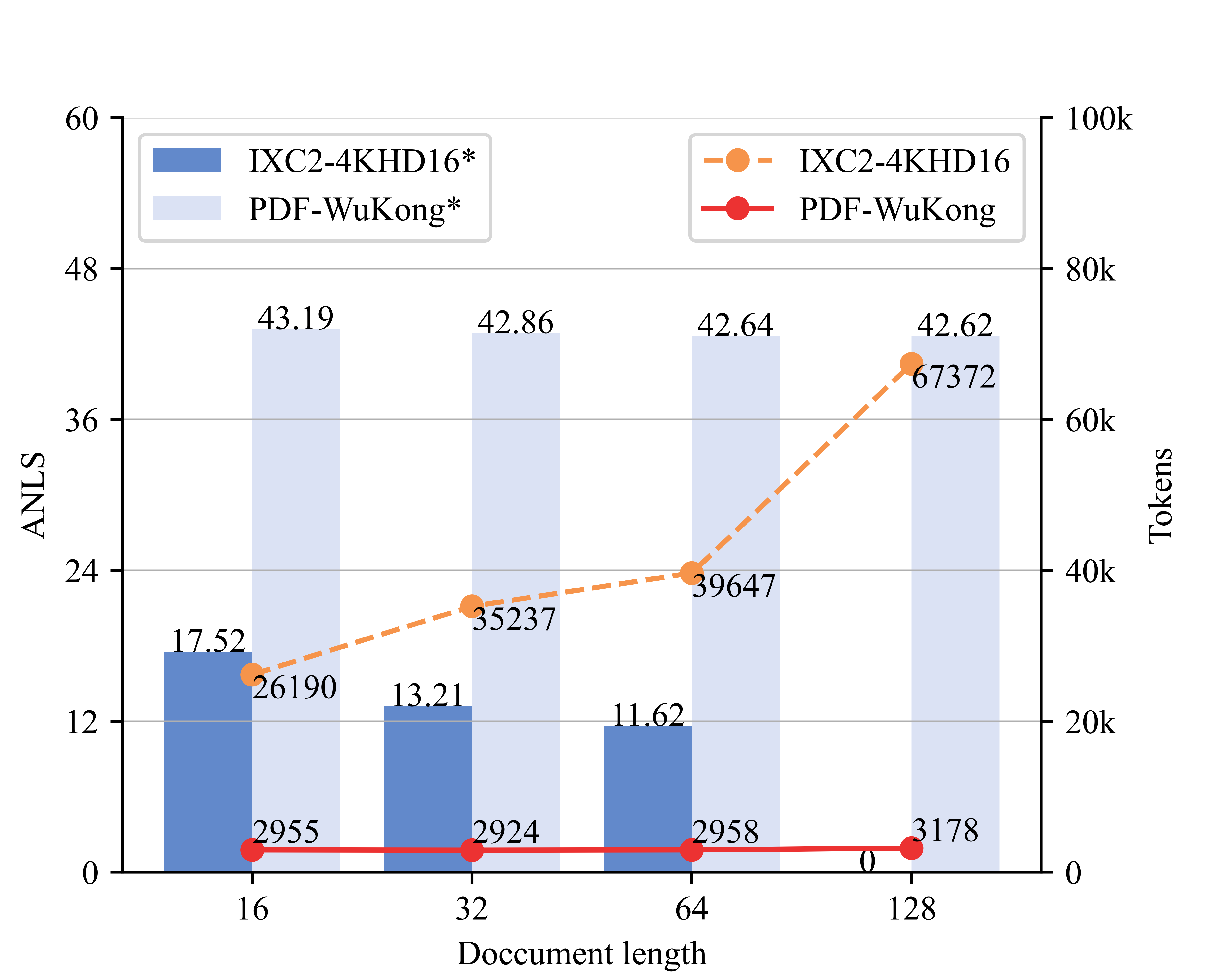}
\caption{Ablation study of different document length}
\label{fig:doc_len}
\end{figure}

To understand the impact of document length on model performance and efficiency, we divide the test set into subsets based on the number of pages per document. Results in Fig.~\ref{fig:doc_len} show that our model's performance and token counts remain stable across documents of varying lengths. This stability indicates that the sparse sampler effectively reduces the input size, regardless of the original document length. 
In contrast, the baseline without the sparse sampler is unable to handle long documents effectively. Its performance deteriorates significantly as the document length increases. The number of tokens also increased dramatically, resulting in huge resource consumption.
These findings highlight the robustness of our model in processing long documents without sacrificing accuracy or incurring extra computational costs.

\setlength{\tabcolsep}{2pt}
\begin{table}[t]
% \small
\caption{\revise{Ablation study of different sampling strategy. P and R, represent the Precision and Recall of the evidence sampling module.}}
% \vspace{-4ex}
\begin{center}
\begin{tabular}{lcccccc}
\toprule
\textbf{Sampling} & \multicolumn{2}{c}{\revise{\textbf{Retrieval}}}&  \multicolumn{3}{c}{\revise{\textbf{QA}}} & \multirow{2}{*}{\textbf{Token}} \\
\cmidrule(lr){2-3}
\cmidrule(lr){4-6}
\textbf{chunks}  &\revise{P}&\revise{R}& \textbf{ANLS} & \textbf{F1} & \textbf{ROUGE}&   \\

\midrule
Top 1 &\revise{46.00}&\revise{45.60}&39.20 &38.28 &35.26&  1186  \\
Top 3 &\revise{22.53}&\revise{66.72}&42.09 &43.06  &39.69& 1452   \\
Top 5 &\revise{14.96}&\revise{73.46}&42.59 & 43.63 & 40.19&1789  \\
Top 10&\revise{8.40}&\revise{81.79}&43.01 & 44.22 & 40.67&2386  \\
Top 15&\revise{5.94}&\revise{86.61}&43.19 &44.57  &42.08&2704   \\
Top 20&\revise{4.63}&\revise{89.64}&43.42 &45.02  &42.30&3364    \\
% GPT-4  & - & - & - & -  \\
\bottomrule
\end{tabular}
 
\label{topn}
\end{center}
\end{table}

\noindent \textbf{Sampling strategy}

%We explore the impact of different numbers of text blocks or diagrams selected by the sparse sampler. As shown in Tab.~\ref{topn}, setting a small top $k$ can lead to missing crucial information needed for accurately answering queries, thus reducing performance. Conversely, a larger top $k$ introduces redundant information and increases computational costs without significantly enhancing accuracy. To strike a balance between performance and resource efficiency, we use the top 5 as the default setting.

\revise{To investigate the effectiveness of the sparse sampler in evidence location, we conduct an ablation study on the number of selected blocks. Beyond standard QA performance evaluation, we further explore the precision and recall of the sampled evidence , following the previous studies~\cite{souibgui2025docvxqa,chen2025towards}. As shown in Tab.~\ref{topn}, setting a small top $k$ results in relatively low recall, which implies that the crucial evidence required for accurately answering queries is often omitted, thereby constraining the model's performance. Conversely, although a larger top-$k$ indeed improves evidence recall, it comes at the cost of a sharp decline in precision, introducing substantial non-evidential redundancy that interferes with the model's reasoning. As a result, when top-$k$ exceeds 5, the model exhibits diminishing returns, where a substantial increase in token consumption yields only marginal gains in QA accuracy. To strike an optimal balance between performance and resource efficiency, we adopt Top-5 as the default setting.}

\section{Conclusion}
We have presented PDF-WuKong, a novel MLLM that effectively addresses the challenges of understanding long PDF documents containing interleaved text and images. By introducing an end-to-end sparse sampling mechanism, our model efficiently extracts the most relevant paragraphs and diagrams in response to user queries, significantly reducing input token size and making the process independent of document length. We also constructed PaperPDF, a bilingual dataset with $1.1M$ question-answer pairs for training and $10k$ pairs for evaluation, specifically tailored for academic PDFs. Experimental results demonstrate that PDF-WuKong not only outperforms existing open-source models but also surpasses proprietary products by an average of 8.6\% in F1 score on the long multimodal PDF understanding task. Our approach maintains high accuracy and efficiency even as document length increases, offering a scalable and interpretable solution for practical applications in document understanding.

\section{Limitation}
\revise{It is important to note that unlike the manually verified test set, the training set is constructed via an automated pipeline. Consequently, it may contain a certain degree of noise stemming from PDF parsing errors (e.g., imperfect layout analysis in complex documents) or alignment inconsistencies in multi-evidence extraction. While this noise is typical in large-scale pre-training datasets, users should be aware of these potential imperfections.}

For our proposed dataset, current documents are mainly limited to academic papers, so the layout format and subject matter of the documents are relatively simple. We will expand the dataset with more diverse documents. Besides, our model is not specifically designed for some global queries, which will become a key research problem in the future.

\section*{Data Availability Statement} 
The datasets used during and/or analysed in the current study are available in our PaperPDF repository \href{https://github.com/yh-hust/PDF-Wukong}{[Link]}, the DocVQA repository \href{https://www.docvqa.org/datasets/docvqa}{[Link]}, the ChartQA repository \href{https://github.com/vis-nlp/ChartQA}{[Link]}, the InfoVQA repository \href{https://www.docvqa.org/datasets/infographicvqa}{[Link]}, the MP-DocVQA repository \href{https://github.com/rubenpt91/MP-DocVQA-Framework}{[Link]}, the DUDE repository \href{https://huggingface.co/datasets/jordyvl/DUDE_loader}{[Link]}, and the MM-NIAH repository \href{https://github.com/OpenGVLab/MM-NIAH}{[Link]}.

\section*{Declaration}
The authors have no relevant financial or non-financial interests to disclose.

\section*{Acknowledgement}
This research is supported by the National Science Fund for Distinguished Young
Scholars of China (Grant No.62225603) and the National Natural Science Foundation of China (Grant No. 62576147).

\bibliographystyle{splncs04}
\bibliography{main}

\clearpage

\begin{appendices}
\onecolumn

\section{Prompts and Examples for PaperPDF Dataset} \label{sec:appendix}

\begin{figure*}[htbp]
    \centering
    \begin{tcolorbox}[colframe=black, colback=white]
    \begin{center}
    \textbf{Cross-paragraph Question Generation Prompt}
    \end{center}
    
    \textbf{Task Definition:} \\
    Based on the selected paragraph from a research paper that share a thematic or conceptual connection, formulate an insightful, open-ended question. This question should reflect the shared themes or concepts of your selections and relate to the broader context of the research paper. \\
    \textbf{Requirements:} 
    \begin{itemize}
        \item Ascertain the underlying connection among the paragraphs and the figures/tables(if provided). 
        \item  Subsequently, craft an insightful, open-ended question that encapsulates the identified themes or connections, aiming to foster analytical thinking and in-depth discussion on the subject matter of the paper. 
        % \item Note that your question should not directly include the ``idx"s of the paragraphs.
    \end{itemize}
    \textbf{Expected Output:}  \\
    \textsc{[Q]:} [Your generated question based on the shared themes or information]
%Here are the selected paragraphs in the paper:
    \end{tcolorbox}
    
    \begin{tcolorbox}[colframe=black, colback=white]
    \begin{center}
    \textbf{Cross-paragraph Answer Generation Prompt}
    \end{center}
    
    \textbf{Task Definition:} \\
    Given some selected paragraphs from a research paper, and ensuring that these paragraphs share a certain level of association, you are to answer a question that is related to the content of these selected paragraphs. \\
    \textbf{Requirements:} \\
    Craft the 2 answers for the question that:
    \begin{itemize}
    \item Are directly derived from the provided figure, excluding information not found within the material.
    \item Are comprehensive and cover all relevant aspects, as presented in the provided figure.
    \end{itemize} 

    \textbf{Expected Output:} \\
    \textsc{[A1]:} [Insert a concise answer here, no more than 20 words.]   \\
    \textsc{[A2]:} [Insert a detailed answer here,including a detailed ``thought chain" or reasoning process.]%, detailing how the conclusions are drawn from the image and caption.] 
    %The question is: 
    \end{tcolorbox}

   \begin{tcolorbox}[colframe=black, colback=white]

    \begin{center}
    \textbf{Cross-paragraph Data Example}
    \end{center}
    
    \textbf{Query:}
    How can we leverage the proposed EDO approach to optimize the selection of datasets for specific algorithms, thereby enhancing the overall performance and validity of the algorithms?
    
    \vspace{5pt}
    \textbf{Text:}\\
    \textbf{Paragraph 7:} Figure 1: On the right: the current path for selecting some algorithm(s), ... \\% based on their validity and performance for a given dataset.On the left: the proposed flip to better understand the space in which `good' datasets exist for an algorithm.\\
    \textbf{Paragraph 17:} Section 2 describes the structure of the proposed method including its parameters and operators.\\
    \textbf{Paragraph 20:} In this section, the details of an algorithm that generates data for which a given function, ... \\ % or (equivalently) algorithm, is well-suited is described.This algorithm is to be referred to as ``Evolutionary Dataset Optimisation" (EDO).  \\
    \begin{minipage}[c]{1\textwidth}
    \vspace{5pt}
    \raggedright
    \textbf{Figure:} Figure 1 \\
    \vspace{5pt}
    \centering
    \includegraphics[width=0.3\textwidth]{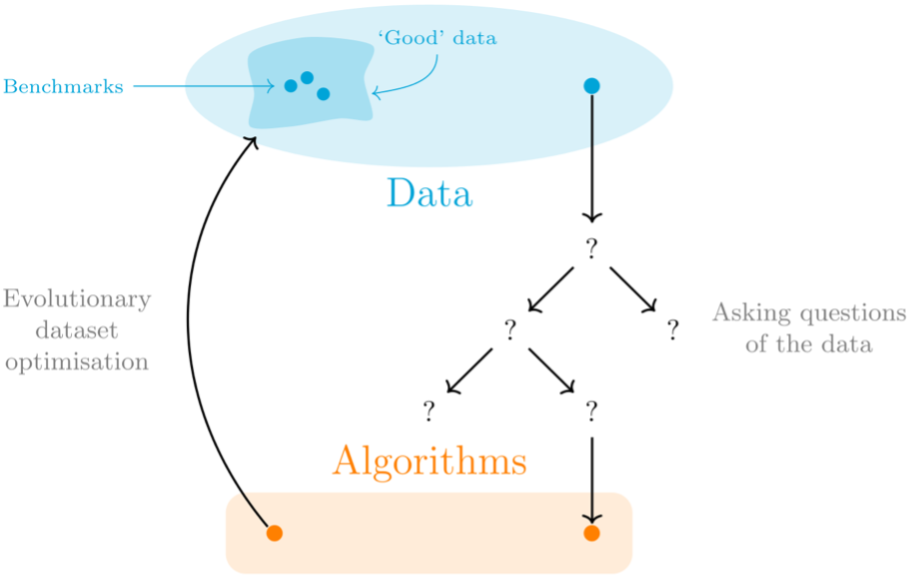}
    
    \end{minipage}
    \hfill
    \begin{minipage}[c]{1\textwidth}
    \vspace{5pt}
    \textbf{Caption:}
    Sample number of rows and columns 2. Sample columns and fill/trim values as needed. \\
    \textbf{Answer 1:}
    The EDO approach identifies optimal datasets, improving algorithm performance and validity through tailored data selection. \\
    \textbf{Answer 2:}
    We can systematically evaluate and select datasets that align closely with the specific requirements of the algorithms in use. This method analyzes various dataset characteristics, ensuring that the chosen data not only matches the algorithm's operational parameters but also enhances its predictive accuracy.
    
    \end{minipage}
    \end{tcolorbox}
    
    \caption{Cross-paragraph Q-E-A triplets generation prompt and data example.}
    \label{fig:cross_para}
\end{figure*}

Fig.~\ref{fig:text_only} - Fig.~\ref{fig:cross_para} show the prompt engineering for constructing the five types of data (\textit{Text-only, Image-only, Image-text, Section, Cross-paragraph}) in single evidence and multiple evidence. There are also corresponding examples for these five types.

% text-only
\begin{figure*}[htbp]
    \centering
    \begin{tcolorbox}[colframe=black, colback=white]
    \begin{center}
    \textbf{Text-only Question Generation Prompt}
    \end{center}
    
    \textbf{Task Definition:} \\
    Create 2 academic questions from a given research paper paragraph.\\
    \textbf{Requirements:}
    Analyze the paragraph thoroughly,understanding its content including the study's objectives, ethods, results,and conclusions.
    Focus on the paragraph,not the entire paper.
    If the paragraph lacks valid information,return `quit'. 
    You should use English. \\
    Develop 2 questions that:
    \begin{itemize}
        \item Are no more than 30 words.
        \item Incorporate knowledge from the paragraph.
        \item Are answered by text instead of one of the multiple choices.
        \item Are elicit detailed responses supported by the text.    \end{itemize} 
    \textbf{Expected Output:} (Return `quit' directly if the paragraph lacks valid information.) \\
    \textsc{[Q1]:}  question1 here \\
    \textsc{[Q2]:} question2 here\\
    \end{tcolorbox}
    
    \begin{tcolorbox}[colframe=black, colback=white]
    \begin{center}
    \textbf{Text-only Answer Generation Prompt}
    \end{center}
    
    \textbf{Task Definition:} \\
    Answer a question based on the material given.  \\
    \textbf{Requirements:} \\
    The answers should:
    \begin{itemize}
        % \item No more than one sentence, less than 20 words.
        \item Be comprehensive and cover all relevant aspects.
        \item Accurately reflect the paragraph's information and insights. 
    \end{itemize} 
You should think step by step and give you answer in the end of your generation like: [thinking procedure]: [A1/A2] \\
    \textbf{Expected Output:} \\
    \textsc{[thinking procedure]:} ... \\
    \textsc{[A1]:} answer1 here, no more than 20 words. \\
    \textsc{[thinking procedure]:} ... \\
    \textsc{[A2]:} answer2 here, no more than one sentence.
    \end{tcolorbox}

%     \begin{tcolorbox}[colframe=black, colback=white]
%     \begin{center}
%     \textbf{Text-only Answer Generation Prompt 2}
%     \end{center}
    
%     \textbf{Task Definition:} \\
%     Answer 2 questions based on the material given.  \\
%     \textbf{Requirements:} \\
%     The answers should be:
%     \begin{itemize}
%         \item Within a few keywords, less than 20 words.
%         \item Comprehensive and cover all relevant aspects.
%         \item Accurately reflect the paragraph's information and insights. 
%     \end{itemize} 
% You should think step by step and give you answer in the end of your generation like: [thinking procedure]: [A1/A2] \\
%     \textbf{Expected Output:} \\
%     \textsc{[thinking procedure]:} ... \\
%     \textsc{[A1]:} answer1 here, within a few keywords. \\
%     \textsc{[thinking procedure]:} ... \\
%     \textsc{[A2]:} answer2 here, within a few keywords.
%     \end{tcolorbox}

    \begin{tcolorbox}[colframe=black, colback=white]
    \begin{center}
    \textbf{Text-only Data Example}
    \end{center}
    
    \textbf{Query:}
    What is the impact of using the same dataset for optimizing and measuring the performance of a model?

    \vspace{10pt}
    \textbf{Text:}
    Here, again, the unfair advantage of optimizing (selecting the models for the ensemble) and measuring performance on the same dataset appears.The advantage is small but systematic for the test split of ISIC (Fig. 5a); it is much more apparent for the challenging collection of clinical images of EDRA Atlas (Fig. 5b). \\

    \textbf{Answer 1:}
    It can lead to an unfair advantage for the model. \\
    \textbf{Answer 2:}
    Optimizing a model involves selecting certain parameters or features that improve its performance on a given dataset. If the same dataset is used to measure the model's performance, it may lead to an unfair advantage as the model has already been ``tuned" to that specific dataset.
    \end{tcolorbox}
    
    \caption{Text-only Q-E-A triplets generation prompt and data example.}
    \label{fig:text_only}
\end{figure*}

% image-only
\begin{figure*}[htbp]
    \centering
    \begin{tcolorbox}[colframe=black, colback=white]
    \begin{center}
    \textbf{Image-only Question Generation Prompt}
    \end{center}
    
    \textbf{Task Definition:} \\
    Formulate 2 academic questions based on a provided figure or table from a research paper.\\
    \textbf{Requirements:}
    The questions must directly reference and integrate information presented in the image and its caption, ensuring a cohesive understanding of the content depicted.
    You should use English. \\
    Develop 2 questions that are:
    \begin{itemize}
        \item No more than 30 words.
        \item Specific to the unique data or details visible in the figures/tables and are answerable only based on the material without inferring or speculating on details not explicitly explained by the figures/tables.
        \item Not mentioning the label of the figure/table directly or use words like `from the figure/table'. 
    \end{itemize} 
    \textbf{Expected Output:}  \\
    \textsc{[Q1]:}  \quad
    \textsc{[Q2]:} 
    \end{tcolorbox}
    
    \begin{tcolorbox}[colframe=black, colback=white]
    \begin{center}
    \textbf{Image-only Answer Generation Prompt}
    \end{center}
    
    \textbf{Task Definition:} \\
    Answer a question based on an image and its caption from a research paper.  \\
    \textbf{Requirements:} \\
    The answers should:
    \begin{itemize}
    % \item No more than one sentence, less than 20 words.
    \item Always use English.
    \item Not infer or speculate on details not explicitly explained by the figures/tables.
    \end{itemize} 
You should think step by step and give you answer in the end of your generation like: [thinking procedure]: [A1/A2] \\
    \textbf{Expected Output:} \\
    \textsc{[thinking procedure]:} ... \quad
    \textsc{[A1]:} answer1 here, no more than 20 words. \\
    \textsc{[thinking procedure]:} ... \quad
    \textsc{[A2]:} answer2 here, no more than one sentence.
    \end{tcolorbox}

%     \begin{tcolorbox}[colframe=black, colback=white]
%     \begin{center}
%     \textbf{Text-image Answer Generation Prompt 2}
%     \end{center}
    
%     \textbf{Task Definition:} \\
%     Answer 2 questions based on the material given.  \\
%     \textbf{Requirements:} \\
%     The answers should be:
%     \begin{itemize}
%         \item Within a few keywords, less than 20 words.
%         \item Comprehensive and cover all relevant aspects.
%         \item Accurately reflect the paragraph's information and insights. 
%     \end{itemize} 
% You should think step by step and give you answer in the end of your generation like: [thinking procedure]: [A1/A2] \\
%     \textbf{Expected Output:} \\
%     \textsc{[thinking procedure]:} ... \\
%     \textsc{[A1]:} answer1 here, within a few keywords. \\
%     \textsc{[thinking procedure]:} ... \\
%     \textsc{[A2]:} answer2 here, within a few keywords.
%     \end{tcolorbox}

\begin{tcolorbox}[colframe=black, colback=white]

    \begin{center}
    \textbf{Image-only Data Example}
    \end{center}
    
    \textbf{Query:}
    Based on Figure 4, which fuzzing technique consistently leads in coverage across all benchmarks over the 24-hour period?
    
    \begin{minipage}[c]{1\textwidth}
    \vspace{5pt}
    \raggedright
    \textbf{Figure:} Figure 4 \\
    \vspace{5pt}
    \centering
    \includegraphics[width=0.5\textwidth]{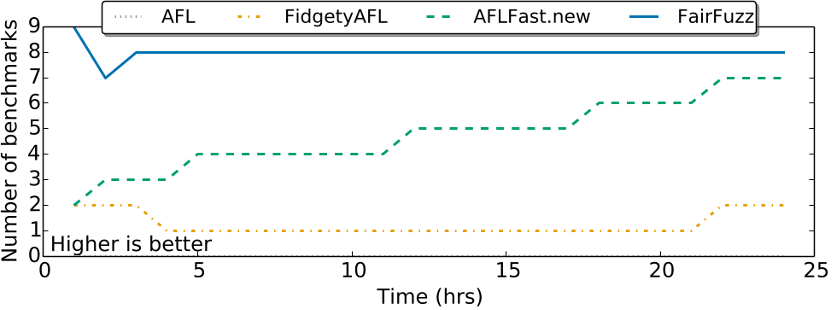}
    
    \end{minipage}
    \hfill
    \begin{minipage}[c]{1\textwidth}
    \vspace{10pt}
    \textbf{Caption:}
    Figure 4: Number of benchmarks on which each technique has the lead in coverage at each hour. A benchmark is counted for multiple techniques if two techniques are tied for the lead. \\

    \textbf{Answer 1:}
    FairFuzz consistently leads in coverage across all benchmarks over the 24-hour period in Figure 4. \\
    \textbf{Answer 2:}
    FairFuzz, highest coverage benchmark count over 24 hours.
    
    \end{minipage}
    
    \end{tcolorbox}
    
    \caption{Image-only Q-E-A triplets generation prompt and data example.}
    \label{fig:train_text_only_pmt}
\end{figure*}

% text-image
\begin{figure*}[htbp]
    \centering
    \begin{tcolorbox}[colframe=black, colback=white]
    \begin{center}
    \textbf{Text-image Question Generation Prompt}
    \end{center}
    
    \textbf{Task Definition:} \\
    Formulate 2 academic questions based on a provided paragraph, a figure or table from a research paper.\\
    \textbf{Requirements:}
    The questions should not rely on the accompanying text but only the figure / table. However, you may use the provided text to understand the figure / table. 
    You should use English. \\
    Develop 2 questions that are:
    \begin{itemize}
        \item No more than 30 words.
        \item Based on a provided figure of a research paper, without relying on accompanying text for the questions. However, you may use the provided text to understand the figure.
        \item Not mentioning the label of the figure/table directly or use words like `from the figure/table'. 
    \end{itemize} 
    \textbf{Expected Output:}  \\
    \textsc{[Q1]:}  \quad
    \textsc{[Q2]:} 
    \end{tcolorbox}
    
    \begin{tcolorbox}[colframe=black, colback=white]
    \begin{center}
    \textbf{Text-image Answer Generation Prompt}
    \end{center}
    
    \textbf{Task Definition:} \\
    Answer a question based on the material given.  \\
    \textbf{Requirements:} \\
    The answers should be:
    \begin{itemize}
    % \item No more than one sentence, less than 20 words.
    \item Always using English.
    \item Comprehensive and cover all relevant aspects, as presented in the provided text and figure. Accurately reflect the information and insights offered by the research paper.

    \end{itemize} 
You should think step by step and give you answer in the end of your generation like: [thinking procedure]: [A1/A2] \\
    \textbf{Expected Output:} \\
    \textsc{[thinking procedure]:} ... \quad
    \textsc{[A1]:} answer1 here, no more than 20 words. \\
    \textsc{[thinking procedure]:} ... \quad
    \textsc{[A2]:} answer2 here, no more than one sentence.
    \end{tcolorbox}

%     \begin{tcolorbox}[colframe=black, colback=white]
%     \begin{center}
%     \textbf{Text-image Answer Generation Prompt 2}
%     \end{center}
    
%     \textbf{Task Definition:} \\
%     Answer 2 questions based on the material given.  \\
%     \textbf{Requirements:} \\
%     The answers should be:
%     \begin{itemize}
%         \item Within a few keywords, less than 20 words.
%         \item Comprehensive and cover all relevant aspects.
%         \item Accurately reflect the paragraph's information and insights. 
%     \end{itemize} 
% You should think step by step and give you answer in the end of your generation like: [thinking procedure]: [A1/A2] \\
%     \textbf{Expected Output:} \\
%     \textsc{[thinking procedure]:} ... \\
%     \textsc{[A1]:} answer1 here, within a few keywords. \\
%     \textsc{[thinking procedure]:} ... \\
%     \textsc{[A2]:} answer2 here, within a few keywords.
%     \end{tcolorbox}

\begin{tcolorbox}[colframe=black, colback=white]

    \begin{center}
    \textbf{Text-image Data Example}
    \end{center}
    
    \textbf{Query:}
    What is the best performing method for both detection and classification, according to the provided figure?
    
    % \vspace{10pt}
    \textbf{Text:}
    3) The last one is our proposed SFCN-OPI with both sibling branches and OPI (Ours in Table 1).
    
    \begin{minipage}[c]{1\textwidth}
    \vspace{5pt}
    \raggedright
    \textbf{Figure:} Table 1 \\
    % \vspace{5pt}
    \centering
    \includegraphics[width=0.35\textwidth]{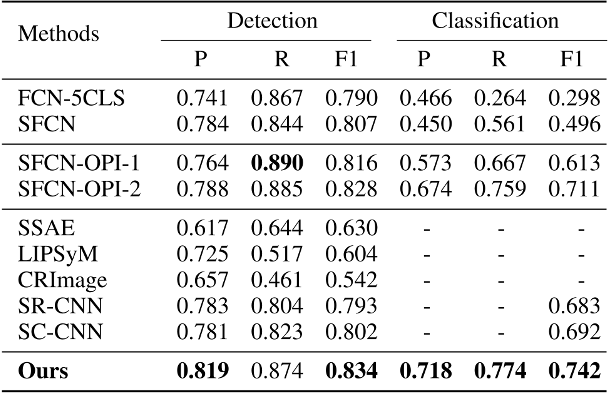}
    
    \end{minipage}
    \hfill
    \begin{minipage}[c]{1\textwidth}
    \vspace{5pt}
    \textbf{Caption:}
    Experimental results of ablation analysis, ... \\

    \textbf{Answer 1:}
    Ours. \\
    \textbf{Answer 2:}
    The best performing method for both detection and classification is Ours, as it achieves the highest F1 scores for both tasks. This can be seen in the ``Ours" row of the table, where the F1 score for detection is 0.834 and the F1 score for classification is 0.742.
    
    \end{minipage}
    \end{tcolorbox}
    
    \caption{Text-image Q-E-A triplets generation prompt and data example.}
    \label{fig:train_text_only_pmt}
\end{figure*}

% sec
\begin{figure*}[htbp]
    \centering
    \begin{tcolorbox}[colframe=black, colback=white]
    \begin{center}
    \textbf{Section Question Generation Prompt}
    \end{center}
    
    \textbf{Task Definition:} \\
    Formulate 2 academic questions based on a section from a research paper.\\
    \textbf{Requirements:} \\
    Carefully read and comprehend the entire provided section of the research paper to ensure a thorough understanding of its content, including key points, findings, methodologies, and conclusions.
    You should Always use English.
    Develop 2 questions that are:
    \begin{itemize}
        \item No more than 30 words.
        \item Requiring an integration of information from all paragraphs and figures/tables in the section.
        \item Not mentioning the label of the figure/table directly or use words like `from the figure/table'. 
        \item Not based on common knowledge or assumptions not supported by the figures and tables.
    \end{itemize}
    \textbf{Expected Output:}  \\
    \textsc{[Q1]:}  \quad
    \textsc{[Q2]:} 
    \end{tcolorbox}
    
    \begin{tcolorbox}[colframe=black, colback=white]
    \begin{center}
    \textbf{Section Answer Generation Prompt}
    \end{center}
    
    \textbf{Task Definition:} \\
    Answer a question based on the material given.  \\
    \textbf{Requirements:} \\
    The answers should be:
    \begin{itemize}
   % \item No more than one sentence, less than 20 words.
    \item Always using English.
    \item Not inferring or speculating on details not explicitly explained by the figures/tables.
    \end{itemize} 
You should think step by step and give you answer in the end of your generation like: [thinking procedure]: [A1/A2] \\
    \textbf{Expected Output:} \\
    \textsc{[thinking procedure]:} ... \quad
    \textsc{[A1]:} answer1 here, no more than 20 words. \\
    \textsc{[thinking procedure]:} ... \quad
    \textsc{[A2]:} answer2 here, no more than one sentence.
    \end{tcolorbox}

%     \begin{tcolorbox}[colframe=black, colback=white]
%     \begin{center}
%     \textbf{Section Answer Generation Prompt 2}
%     \end{center}
    
%     \textbf{Task Definition:} \\
%     Answer 2 questions based on the material given.  \\
%     \textbf{Requirements:} \\
%     The answers should be:
%     \begin{itemize}
%         \item Within a few keywords, less than 20 words.
%         \item Comprehensive and cover all relevant aspects.
%         \item Accurately reflect the paragraph's information and insights. 
%     \end{itemize} 
% You should think step by step and give you answer in the end of your generation like: [thinking procedure]: [A1/A2] \\
%     \textbf{Expected Output:} \\
%     \textsc{[thinking procedure]:} ... \\
%     \textsc{[A1]:} answer1 here, within a few keywords. \\
%     \textsc{[thinking procedure]:} ... \\
%     \textsc{[A2]:} answer2 here, within a few keywords.
%     \end{tcolorbox}
 \begin{tcolorbox}[colframe=black, colback=white]

    \begin{center}
    \textbf{Section Data Example}
    \end{center}
    
    \textbf{Query:}
    How does the qualitative evaluation of extractive summarizers using word clouds elucidate the differences in content focus between the original documents and the summaries?
    
    \vspace{5pt}
    \textbf{Text:}\\
    Here we use word cloud representations to give an intuitive interpretation of the content in the generated extractive summarizers... 
    
    Figure 3 shows a word cloud made by the aggregation of all the summaries generated by the PKUSUMSUM-Centroid method...

    The images clearly show a contrast of content... 

    \begin{minipage}[c]{1\textwidth}
    \vspace{5pt}
    \raggedright
    \textbf{Figure:} Figure 3 \\
    \vspace{5pt}
    \centering
    \includegraphics[width=0.3\textwidth]{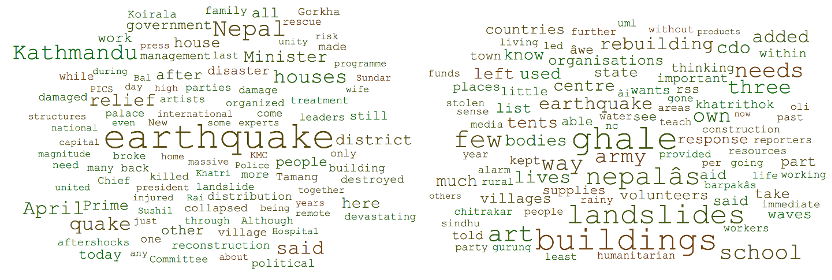}
    
    \end{minipage}
    \hfill
    \begin{minipage}[c]{1\textwidth}
    \vspace{5pt}
    \textbf{Caption:}
    Figure 3: The word clouds representing summaries generated by PKUSUMSUM-Centroid method (left) and original documents without the content of those summaries (right). \\
    \textbf{Answer 1:}
    Visual contrast in word frequency highlights content focus differences.\\
    \textbf{Answer 2:}
    Word clouds highlight the prominent themes in summaries versus original texts by displaying relative word frequencies visually. 
   
    \end{minipage}
    \end{tcolorbox}
    
    \caption{Section Q-E-A triplets generation prompt and data example.}
    \label{fig:train_text_only_pmt}
\end{figure*}

% this is the prompt for the training set.
%cross-para

\end{appendices}

\end{document}